\documentclass[a4paper, 11pt]{article}

\usepackage{authblk}
\usepackage[colorlinks]{hyperref}
\usepackage{subfig}
\usepackage[title]{appendix}
\usepackage{multirow}
\usepackage{algorithm,algcompatible}
\usepackage{algpseudocode}
\usepackage[utf8]{inputenc}
\usepackage[T1]{fontenc}
\usepackage{textcomp}
\usepackage{subfig}
\usepackage{pgf}
\usepackage{makecell}
\usepackage{tablefootnote}
\usepackage[T1]{fontenc}
\usepackage{natbib}
\usepackage{hyperref}
\usepackage{amsmath}
\usepackage{amsfonts}
\usepackage{lipsum}
\usepackage{amsxtra}
\usepackage{amssymb}
\usepackage{amsmath} 
\usepackage{bbm}
\usepackage{amsfonts}
\usepackage{float}
\usepackage[english]{babel}
\usepackage{amsthm}
\usepackage[left=2.05cm,
			right=2.05cm,
			top=2.05cm,
			bottom=2.05cm,]{geometry}

\hypersetup{
    colorlinks=true,
    linkcolor=red,
    citecolor=cyan,
}

\def\mub{\boldsymbol{\mu}}
\def\Sigmab{\boldsymbol{\Sigma}}
\def\xb{\boldsymbol{x}}
\def\zb{\boldsymbol{z}}
\def\thetab{\boldsymbol{\theta}}

\newtheorem{prop}{Proposition}

\newtheorem{remark}{Remark}

\begin{document}
\title{A Robust and Flexible EM Algorithm for Mixtures of Elliptical Distributions with Missing Data}
\date{\textbf{This is a preprint version of the research paper published in \textit{IEEE Transaction on Signal Processing}} : F. Mouret, A. Hippert-Ferrer, F. Pascal and J. -Y. Tourneret, "A Robust and Flexible EM Algorithm for Mixtures of Elliptical Distributions with Missing Data," in IEEE Transactions on Signal Processing, vol. 71, pp. 1669-1682, 2023, doi: \href{https://doi.org/10.1109/TSP.2023.3267994}{10.1109/TSP.2023.3267994}.}

\author[1,2]{Florian Mouret}
\author[3]{Alexandre Hippert-Ferrer}
\author[3]{Frédéric Pascal}
\author[2]{Jean-Yves Tourneret}
\affil[1]{TerraNIS, Ramonville-Saint-Agne, 31520, France}
\affil[2]{University of Toulouse, IRIT-INP-ENSEEIHT / TéSA, Toulouse, 31000, France}
\affil[3]{Université Paris-Saclay, CNRS, CentraleSupélec, Laboratoire des signaux et systèmes, Gif-sur-Yvette, 91190, France}

\maketitle

\abstract{This paper tackles the problem of missing data imputation for noisy and non-Gaussian data. A classical imputation method, the Expectation Maximization (EM) algorithm for Gaussian mixture models, has shown interesting properties when compared to other popular approaches such as those based on $k$-nearest neighbors or on multiple imputations by chained equations. However, Gaussian mixture models are known to be non-robust to heterogeneous data, which can lead to poor estimation performance when the data is contaminated by outliers or follows non-Gaussian distributions. To overcome this issue, a new EM algorithm is investigated for mixtures of elliptical distributions with the property of handling potential missing data. This paper shows that this problem reduces to the estimation of a mixture of Angular Gaussian distributions under generic assumptions (\textit{i.e.}, each sample is drawn from a mixture of elliptical distributions, which is possibly different for one sample to another). In that case, the complete-data likelihood associated with mixtures of elliptical distributions is well adapted to the EM framework with missing data thanks to its conditional distribution, which is shown to be a multivariate $t$-distribution. Experimental results on synthetic data demonstrate that the proposed algorithm is robust to outliers and can be used with non-Gaussian data. Furthermore, experiments conducted on real-world datasets show that this algorithm is very competitive when compared to other classical imputation methods.}


\maketitle

\section{Introduction}\label{sec:intro}

Missing data is a recurrent problem in data analysis that has been studied for decades \citep{Anderson_1957, DEMP1977, little:rubin:2002, van2018flexible}. Missing data appear in a wide range of applications including biomedical signal processing, medical imaging \citep{CISMONDI201363, Mirza_2019} and remote sensing \citep{Shen2015}. As an example, in remote sensing applications, missing data can be due to acquisition problems, cloud coverage or poor atmospheric conditions. The missing data problem is also of critical importance in applications relying on techniques that are non-robust to the absence of data, which is often the case with classical machine learning approaches (\textit{e.g.}, most of the regression or classification algorithms provided in the benchmark Python library \textit{scikit-learn} \citep{scikit-learn} cannot be used with missing data). Moreover, having access to imputed values can be interesting for the end-user. An example that will be investigated in this paper is crop monitoring based on remote sensing images, which requires to have access to timely and accurate information on the crop status \citep{MORAN1997319, Mouret_2021_gmm}.

\subsection{Related work}

Missing value imputation (MVI) is a common solution to bypass the data incompleteness. Two main approaches are generally used for this task, namely \textit{statistical} and \textit{machine learning} techniques \citep{Lin_2020}. Some imputation strategies are very simple, \textit{e.g.}, imputing the missing data by the mean or mode of the feature or by linear interpolation when working with time series. The simplicity of these methods and their straightforward implementation have motivated their use in various applications \citep{Farhangfar_2007}. However, their performance can be limited in some practical applications, motivating the use of more sophisticated techniques for MVI. Methods based on the Expectation Maximization (EM) algorithm have been widely used for MVI \citep{Lin_2020}. As explained in \citep{Ghahramani_1994}, the EM algorithm can be naturally extended to handle missing data, the problem of mixture estimation being itself a missing data problem. Similarly to clustering or classification tasks, a particular attention has been devoted to the EM algorithm for Gaussian Mixture Models (GMM) with missing data \citep{DEMP1977, Ghahramani_1994, EIROLA201432}. However, GMM estimation is known to be non-robust to noisy data and outliers \citep{Campbell_1984,Tadjudin_2000, roizman2020flexible}. Moreover, when the data has a non-Gaussian distribution (\textit{i.e.}, with heavier or lighter tails than the Gaussian distribution), the performance of GMM estimation algorithms might decrease significantly  \citep{Fraley_2002}. In the complete-data case, various strategies have been investigated to solve these issues ranging from robust parameter estimation to the use of non-Gaussian distributions such as multivariate $t$- or hyperbolic distributions \citep{Campbell_1984, Tadjudin_2000, Peel_2000, Browne_2015}. Some of these strategies have been adapted to the missing data case, \textit{e.g.}, using multivariate $t$-distributions \cite{WANG2004701} or skew $t$-distributions \cite{WEI201918}. However, their extension to the missing data case is generally not straightforward since new conditional expectations have to be computed during the expectation step of the EM algorithm. 

\subsection{Contributions of this work}

A flexible EM algorithm (FEM) was recently investigated in \cite{roizman2020flexible, roizman2021_eusipco}, showing good properties for the clustering of noisy and non-Gaussian data. An outstanding property of this algorithm is its robustness to the underlying data distribution when assuming cluster-independent density generators, see \cite[Proposition 4]{roizman2020flexible}. This property was used to build a versatile EM clustering algorithm characterized by a simple parameter tuning (\textit{i.e.}, self-contained).

This paper proposes to extend the FEM algorithm to handle missing data. The resulting algorithm is able to perform an efficient MVI, which generally outperforms the classical EM for GMMs\footnote{The proposed approach could also be used for clustering, as in the case without missing data. However, this task is not considered in this paper.}. The main contributions of this work can be summarized as follows:

\begin{itemize}
    \item A new EM algorithm for mixtures of elliptical distributions potentially affected by missing data is derived. The algorithm assumes that each sample is drawn from a mixture of elliptical distributions, which is possibly different for one sample to another. Under these generic assumptions, the complete likelihood is shown to be upper bounded by the likelihood of a mixture of Angular Gaussian (AG) distributions. Moreover, the conditional distribution of the complete likelihood used in the considered EM framework is shown to be a Student's $t$-distribution.
    \item Algorithms for the implementation of the proposed EM algorithm are provided. In addition, we show that the proposed algorithm is \textit{intuitive} in the sense that its derivation is very similar to the EM for GMM in the missing data case.
    \item Imputation results obtained on different synthetic and real world datasets are presented to evaluate the flexibility of the proposed approach.
\end{itemize}

\subsection{Notations}

In the following, $a$ refers to a scalar quantity, $\boldsymbol{a}$ to a vector and $\boldsymbol{A}$ to a matrix. The notation $\text{det}(\boldsymbol{A})$ (resp. $\text{tr}(\boldsymbol{A})$) refers to the determinant of matrix $\boldsymbol{A}$ (resp. trace of matrix $\boldsymbol{A}$). Moreover $\boldsymbol{A}^T$ is the transpose of $\boldsymbol{A}$.

\bigskip
The rest of this paper is organized as follows. \autoref{sec:background} presents the general context and background necessary to understand the FEM method. \autoref{sec:method} derives the proposed algorithm for missing data with appropriate theoretical justifications. \autoref{sec:results} evaluates the performance of the proposed FEM algorithm to impute missing values in various datasets, using both synthetic and real world data. Note that this paper focuses on the data reconstruction task, the performance of the FEM algorithm for clustering was evaluated in \cite{roizman2020flexible} in the complete data case. \autoref{sec:conclusion} finally draws some conclusions and presents some future work that would deserve to be conducted.

\section{Background: an EM algorithm for mixtures of elliptical distributions}\label{sec:background}

Elliptical Symmetric (ES) distributions refer to a broad family of distributions generalizing the multivariate normal distribution, such as the famous multivariate $t$-distribution \citep{Peel_2000} or the multivariate generalized Gaussian distributions \citep{Pascal_2013}. They have been used in a wide range of applications to deal with heavy-tailed distribution or outliers \citep{Conte_2002, Tadjudin_2000}. Their flexibility and robustness have proven to be very interesting in tasks such as classification or clustering when compared to standard methods \citep{HIPPERTFERRER2022108460, roizman2020flexible}, particularly for heterogeneous or noisy datasets. More information regarding elliptical distributions can be found in the pioneering paper from \cite{Kelker_1970}, which introduced for the first time this generalization of the Gaussian distribution. A more recent survey in the complex case can be found in \cite{Ollila_2012}.

This section briefly recalls the FEM algorithm in its standard formulation, \textit{i.e.}, without missing data, as proposed in \cite{roizman2020flexible} (for more details and justifications, the reader is invited to consult this reference). We focus on the case where the density generator is the same for each component, which allows us to derive generic estimators that do not depend on the underlying distribution of the data. In a second step, we extend this procedure to the missing data scenario, which is the main contribution of this work. The resulting algorithm is very intuitive and has the interesting property to be robust to outliers. 

\subsection{Data model and complete log-likelihood}

Suppose that each sample $\xb_i \in \mathbb{R}^m$ of the dataset $\mathcal{X}=  \{ \xb_1,...,\xb_N \}$ (containing $N$ samples of dimension $m$) is drawn from a mixture of distributions with the following probability density function (pdf):
\begin{equation}
 f_{i, \thetab}(\xb_i) = \sum_{k=1}^{K} \pi_k f_{i, \thetab_k} (\xb_i) \ \text{with} \ \sum_{k=1}^{K} \pi_k = 1,
\end{equation}
where $\pi_k$ denotes the \textit{a priori} probability of class $k$, $\thetab=(\thetab_1^T,...,\thetab_K^T)^T$ and $f_{i, {\thetab_k}}$ is the pdf of $\xb_i$ (that is potentially different for each sample). Note that the parameters of cluster $k$ are grouped into the vector $\thetab_k$. This paper assumes that $f_{i, \thetab_k}$ is an ES distribution with mean vector $\mub_k$ and covariance matrix $\tau_{ik} \Sigmab_k$ \citep{Kelker_1970}, whose pdf can be written:
\begin{equation}\label{eq:f_standard}
    f_{i, \thetab_k}(\xb_i) = A_{ik} \det(\Sigmab_k)^{-1/2} \tau_{ik}^{-m/2}g_{i,k}\left(\frac{(\xb_i - \mub_k)^T \Sigmab_k^{-1}(\xb_i - \mub_k)}{\tau_{ik}} \right),
\end{equation}
where $A_{ik}$ is a normalization constant and $g_{i,k}$ is a density generator such that \autoref{eq:f_standard} defines a pdf. $\Sigmab_k$ is referred to as the scatter matrix, which defines the structure of the covariance of $\xb_i$ (in particular $\Sigmab_k$ is equal to the covariance matrix of $\xb_i$ up to a scale factor). Finally, $\tau_{ik}$ is known as the scale or nuisance parameter and is not of direct interest when estimating the other model parameters. After introducing the scale factor
$$
s_{ik} = \frac{(\xb_i - \mub_k)^T \Sigmab_k^{-1}(\xb_i - \mub_k)}{\tau_{ik}},
$$
the pdf $f_{i, \thetab}$ can be expressed as follows:
\begin{equation}\label{eq:f_ibis}
    f_{i, \thetab_k}(\xb_i) = A_{ik} \det(\Sigmab_k)^{-1/2} \left[(\xb_i - \mub_k)^T \Sigmab_k^{-1}(\xb_i - \mub_k)\right]^{-m/2} s_{ik}^{m/2}g_{i,k}(s_{ik}).
\end{equation}

In the rest of this paper, \textbf{we will suppose that the density generator is the same for each component} (but is possibly different from one sample to another), \textit{i.e.}, $g_{i,k} = g_i$, leading to
\begin{equation}\label{eq:f_iter}
    f_{i, \thetab_k}(\xb_i) = A_{i} \det(\Sigmab_k)^{-1/2} \left[(\xb_i - \mub_k)^T \Sigmab_k^{-1}(\xb_i - \mub_k)\right]^{-m/2} s_{ik}^{m/2}g_{i}(s_{ik}).
\end{equation}

In this paper, we show that this specific structure for the density generator allows us to have a model that is as generic as possible with intuitive and relatively simple derivations. Indeed, assuming the pdf depends on sample $i$ is more generic without any complexity added to the algorithm. Moreover, while preserving the generality of the estimated model, this assumption on the density generator provides an algorithm allowing the model parameters to be estimated without knowing the generator $g_i$, which is an interesting property of the FEM algorithm \cite{roizman2020flexible}. Note that a precise knowledge of the underlying data distribution could improve the estimation of the model parameters \citep{roizman2020flexible}. However, the main interest of the FEM algorithm presented in this paper is that it can be used without any a priori on the data distribution (\textit{e.g.}, if the $\xb$ observations are identically distributed or not).

The complete log-likelihood for a mixture of elliptical distributions can be defined by introducing latent vectors
$\mathcal{Z}= \{ \zb_1,...,\zb_N \}$ containing the cluster labels for the different observed vectors. More precisely, for each sample $\xb_i$, the latent vector $\zb_i=(z_{i1},...,z_{iK})^T$ is such that $z_{ik}= 1$ if the vector $\xb_i$ belongs to the $k$th component of the model and $z_{ik}= 0$ otherwise: 
\begin{equation}\label{eq:Lc_comple_data}
    \log \mathcal{L}_c(\thetab; \mathcal{X}, \mathcal{Z}) = \sum_{i=1}^{N} \sum_{k=1}^{K} z_{ik} \log\left( \pi_k f_{i, \thetab_k} (\xb_i)\right).
\end{equation}
Based on \autoref{eq:f_iter}, $\log \mathcal{L}_c(\thetab; \mathcal{X}, \mathcal{Z})$ can be rewritten as:
\begin{equation}\label{eq:Lc_comple_data_bis}
\begin{split}
    \log \mathcal{L}_c(\thetab; \mathcal{X}, \mathcal{Z}) = \sum_{i=1}^{N} \sum_{k=1}^{K} z_{ik} \left[ \ell_{0k}(\xb_i; \pi_k, \mub_k, \Sigmab_k) + \ell_{ik}(\xb_i ; \pi_k, \mub_k, \Sigmab_k, \tau_{ik}) \right]
\end{split}
\end{equation}
with

\begin{align}
    \begin{split}
    \ell_{0k}(\xb_i ; \pi_k, \mub_k, \Sigmab_k) = & \log(\pi_k) + \log(A_{i}) + \frac{1}{2} \log\left(\det(\Sigmab_k^{-1})\right) \\ & - \frac{m}{2} \log\left((\xb_i - \mub_k)^T \Sigmab_k^{-1}(\xb_i - \mub_k)\right), \end{split} \label{eq:l0} \\
    \ell_{ik}(\xb_i ; \pi_k, \mub_k, \Sigmab_k, \tau_{ik}) = & \log \left(s_{ik}^{m/2}g_{i}(s_{ik})\right).
\end{align}

\subsection{The M-step}\label{sec:M_step_complete}

Using the EM algorithm, where $t$ denotes the current iteration, we can derive the new set of parameters $\thetab^{(t+1)}$ based on the current set of parameters $\thetab^{(t)}$ and $p_{ik} = E[z_{ik} \vert \xb_i, \thetab^{(t)}]$, the probability that sample $i$ has been generated by component $k$. The probabilities $p_{ik}$ (also referred to as responsibilities) are computed in the E-step defined in the next section. We begin by the M-step since interesting results regarding the estimation of $\tau_{ik}$ will allow the E-step to be simplified significantly. For brevity, denote as $\thetab^{(t)} = \thetab$, \textit{i.e.}, $\mub_k^{(t)} = \mub_k$, $\Sigmab_k^{(t)} = \Sigmab_k$, $\tau_{ik}^{(t)} = \tau_{ik}$  and $\pi_k^{(t)} = \pi_k$ the current set of parameters.
\begin{itemize}
    \item \textbf{Estimation of the nuisance parameters $\tau_{ik}$}: One can observe that $\tau_{ik}$ is only related to the term $\ell_{ik}(\xb_i ; \pi_k, \mub_k, \Sigmab_k, \tau_{ik})$ of the complete log-likelihood. For fixed $(\pi_k, \mub_k, \Sigmab_k)$, the value of $\tau_{ik}$ maximizing $p_{ik}\log(s_{ik}^{m/2}g_{i}(s_{ik}))$ is:
    \begin{equation}\label{eq:result_tau_1}
        \tau_{ik} = \frac{(\xb_i - \mub_k)^T \Sigmab_k^{-1}(\xb_i - \mub_k)}{a_{im}},
    \end{equation}
    with $a_{im} = \text{arg sup}_t (t^{m/2} g_i(t))$ (see proof and complementary results in \cite{roizman2020flexible}, in particular, it has been shown that $a_{im}\sim m$ when $m$ is sufficiently large). After replacing $\tau_{ik}$ by its estimate in $s_{ik}$, the following results can be obtained:
    \begin{equation}\label{eq:result_tau}
        s_{ik} = \frac{(\xb_i - \mub_k)^T \Sigmab_k^{-1}(\xb_i - \mub_k)}{\tau_{ik}} = a_{im},
    \end{equation}
    \begin{equation}
     \ell_{ik}(\xb_i ; \pi_k, \mub_k, \Sigmab_k, \tau_{ik}) = \log(s_{ik}^{m/2}g_{i}(s_{ik})) = \log(a_{im}^{m/2} g_i(a_{im})).
    \end{equation}
    This central result indicates that $\ell_{ik}( \xb_i ; \pi_k, \mub_k, \Sigmab_k, \tau_{ik})$ does not depend on $(\pi_k, \mub_k, \Sigmab_k)$ within the EM framework, and thus only depends on $a_{im}$. Consequently, estimating these parameters using the complete log-likelihood can be done using $\ell_{0k}(\xb_i ; \pi_k, \mub_k, \Sigmab_k)$ only, which reduces to estimate the parameters of a mixture of AG distributions. More formally, we denote as $AG_{m}(\mub_k, \Sigmab_k)$ the AG distribution studied in \citep{Ollila_2012}, with pdf 
    \begin{equation}\label{eq:pdf_AG_distribution}
    f_{AG}(\xb_i)~=~B_{i} \det(\Sigmab_k)^{-1/2} \left[(\xb_i - \mub_k)^T \Sigmab_k^{-1}(\xb_i - \mub_k)\right]^{-m/2},    
    \end{equation}
    where $B_i$ is a normalization constant.
    
    \begin{remark}
    \textit{The estimation of $\tau_{ik}$ reduces $\Sigmab_k^{-1/2}(\xb_i - \mub_k)$ to lie on a hypersphere, which explains the apparition of the AG distribution. Indeed, the stochastic representation theorem (see \textit{e.g., \cite{roizman2020flexible}}) for $\xb_i$ distributed according to an elliptical distribution can be stated as follows: \begin{equation*}
        \xb_i \overset{d}{=} \mub_k + \sqrt{\mathcal{Q}_{ik}} \mathbf{A}_k \mathbf{u}_i
    \end{equation*}  where $\overset{d}{=}$ stands for ``is distributed as'', $\mathcal{Q}_{ik}$ is a positive random variable independent of $\mathbf{u}_i$, $ \mathbf{A}_k$ is such that $ \mathbf{A}_k \mathbf{A}_k^T=\Sigmab_k$ and $\mathbf{u}_i$ is a uniform random vector on the unit hypersphere. Since $\mathcal{Q}_{ik}$ is distributed as $(\xb_i - \mub_k)^T \Sigmab_k^{-1}(\xb_i - \mub_k)$ \citep{Ollila_2012}, the following result is obtained:
    \begin{equation*}
    \cfrac{\Sigmab_k^{-1/2}(\xb_i -\mub_k)}{\sqrt{\mathcal{Q}_{ik}}} \overset{d}{=} \mathbf{u}_i,       \end{equation*}
    }
    which shows that the normalized observation $\xb_i$ lies on the unit hypersphere.
    \end{remark}
    
    \bigskip
    \item \textbf{Estimating the model parameters $(\pi_k, \mub_k, \Sigmab_k)$}: maximizing the log-likelihood with respect to the model parameters leads to the following expressions defined through fixed-point equations:
    
    \begin{align}
        \pi_k= & \frac{1}{N} \sum_{i=1}^N p_{ik},
        \\
        \mub_k= & \sum_{i=1}^{N} \frac{w_{ik}p_{ik} \xb_i}{\sum_{l=1}^N w_{lk}p_{lk}}, \label{eq:mean_FEM}
        \\
        \Sigmab_k = & m \sum_{i=1}^{N} \frac{p_{ik} w_{ik} (\xb_i - \mub_k)(\xb_i - \mub_k)^T}{\sum_{l=1}^N p_{lk}}, \label{eq:cov_FEM}
    \end{align}
    
where $w_{ik} = \frac{1}{(\xb_i - \mub_k)^T \Sigmab_k^{-1}(\xb_i - \mub_k)}$. One can notice that \autoref{eq:mean_FEM} and \autoref{eq:cov_FEM} are classical expressions of the mean and covariance matrix in a robust estimation framework. More precisely, $w_{ik}$ are weights reducing the influence of outlier samples (see for instance \citep{Campbell_1984, Tadjudin_2000}, where similar forms are obtained using the Huber function for the weight function). Typically, when $w_{ik}$ is close to zero, the considered sample will have little influence on the estimation of the model parameters.

Finally, it should be noted that the scatter matrix $\Sigmab_k$ is equal to the covariance matrix up to a scale factor. In this work, we choose to fix the trace of $\Sigmab_k$ to $m$, as in \cite{roizman2020flexible}. The aforementioned reference has shown that the scale of $\Sigmab_k$ has no influence on the clustering results. In \autoref{sec:M_step_missing}, we will show that the scale of $\Sigmab_k$ has also no influence either on the imputation results in the missing data case, which is an important result.
\end{itemize}

\subsection{The E-step}

In the E-step, one needs to compute $E[\log \mathcal{L}_c(\thetab; \mathcal{X}, \mathcal{Z})) \vert \thetab^{(t)}]$, which reduces to evaluate $\hat{p}_{ik} = E[z_{ik} \vert \xb_i, \thetab^{(t)}]$ when there is no missing data. By replacing unknown parameters by the current state of parameters at iteration $t$, \textit{i.e.}, by $\thetab^{(t)}$, $\hat{p}_{ik}$ can be computed as follows:
\begin{equation}
    \hat{p}_{ik} = \frac{\pi_k f_{i, \thetab_k}(\xb_i)}{\sum_{j=1}^{K} \pi_j f_{i, \thetab_j}(\xb_i)}.
\end{equation}
Using the result obtained in \autoref{eq:result_tau} leads to:
\begin{equation}\label{eq:e_step_standard}
    \hat{p}_{ik} = \frac{\pi_k \det(\Sigmab_k)^{-1/2} \left[(\xb_i - \mub_k)^T \Sigmab_k^{-1}(\xb_i - \mub_k)\right]^{-m/2}}{\sum_{j=1}^{K} \pi_j \det(\Sigmab_j)^{-1/2} \left[(\xb_i - \mub_j)^T \Sigmab_k^{-1}(\xb_i - \mub_j)\right]^{-m/2}}.
\end{equation}
It is important to note here that the result obtained in \autoref{eq:e_step_standard} shows that $\hat{p}_{ik}$ does not depend on the density generator $g$ when $g_{ik} = g_i$. This result is particularly important since knowing the precise data distribution (and the corresponding density generator) is often not possible in practical applications. 

\subsection{Algorithm}

Algorithm~\ref{alg:cap} provides the pseudo-code for the FEM implementation in the complete data case. The reader is invited to consult \cite{roizman2020flexible} for a discussion regarding implementation details and numerical considerations. Regarding the initialization of $\thetab^{(0)}$, it is possible to use random values. However, a more efficient way is to use a fast clustering algorithm (such as the K-means algorithm) to have a more relevant initial guess for the set of parameters. This is for instance the strategy adopted in the Python library \textit{scikit-learn} for the estimation of GMM.

\begin{algorithm}[H]
\footnotesize
\caption{Scheme of the FEM algorithm in the complete data case \citep{roizman2020flexible}.}\label{alg:cap}
\textbf{Input:} Data  $\{\textbf{\textit{x}}_i\}_{i=1}^n$, $K$ the number of clusters \\
\textbf{Output:} Clustering labels $\mathcal{Z}=\{z_i\}_{i=1}^n$ and model parameters $\thetab =$ $ \{\pi_1, ..., \pi_k, $ $\boldsymbol{\mu}_1, ..., \boldsymbol{\mu}_k, \boldsymbol{\Sigma}_1, ..., \boldsymbol{\Sigma}_K \}$
\begin{algorithmic}[1]
\State Initialize $\thetab^{(0)}$ (randomly or using a clustering algorithm);
\State $t \leftarrow 1$;

\While{not convergence}

\For{$1 \leq k \leq K$:}\Comment{\textbf{E-step}}
\State $ p_{ik}^{(t)} = \frac{\pi_k^{(t-1)} \det(\Sigmab_k^{(t-1)})^{-1/2} \left[(\xb_i - \mub_k^{(t-1)})^T (\Sigmab_k^{(t-1)})^{-1}(\xb_i - \mub_k^{(t-1)})\right]^{-m/2}}{\sum_{j=1}^{K} \pi_j^{(t-1)} \det(\Sigmab_j^{(t-1)})^{-1/2} \left[(\xb_i - \mub_j^{(t-1)})^T (\Sigmab_k^{(t-1)})^{-1}(\xb_i - \mub_j^{(t-1)})\right]^{-m/2}};$
\EndFor

\For{$1 \leq k \leq K$:}\Comment{\textbf{M-step}}
\medskip
\State $\pi_{k}^{(t)} = \frac{1}{N} \sum_{i=1}^N p_{ik}^{(t)}$;
\medskip
\State set $\mub_k^{'} = \mub_k^{(t-1)}$ and $\Sigmab_k^{'} = \Sigmab_k^{(t-1)}$;
    \medskip
    \While{not convergence}\Comment{Fixed-point loop}
    \State $w_{ik} = \frac{1}{(\xb_i - \mub_k^{'})^T (\Sigmab_k^{'})^{-1}(\xb_i - \mub_k^{'})};$
    \medskip
    \State $\mub_k^{''} = \sum_{i=1}^{N} \frac{w_{ik}p_{ik}^{(t)} \xb_i}{\sum_{l=1}^N w_{lk}p_{lk}^{(t)}};$
    \medskip
    \State $ \Sigmab_k^{''} = m \sum_{i=1}^{N} \frac{w_{ik} p_{ik}^{(t)} (\xb_i - \mub_k^{'})(\xb_i - \mub_k^{'})^T}{\sum_{l=1}^N p_{lk}^{(t)}};$
    \medskip
    \State Update: $\mub_k^{'} = \mub_k^{''}$ and $\Sigmab_k^{'} = \Sigmab_k^{''}$;
    \EndWhile
    \State Update: $\mub_k^{(t)} = \mub_k^{''}$ and $\Sigmab_k^{(t)} = \Sigmab_k^{''}$;
\EndFor

\State $t \leftarrow t+1 $;

\EndWhile
\State Set $z_i$ as the index $k$ that has the maximum $p_{ik}$;
\end{algorithmic}
\end{algorithm}

\newpage
\section{A generalization to incomplete data}\label{sec:method}

This section explains how to extend the previous FEM algorithm to handle missing data. In this case, each sample can be decomposed into $\xb_i = (\xb_i^{o_i}, \xb_i^{m_i})$, where $\xb_i^{o_i}$ and $\xb_i^{m_i}$ are the vectors of observed and missing variables respectively (we denote $\mathcal{X}^o$ the sets of all observed variables and $\mathcal{X}^m$ the set of all missing variables). More generally, the superscripts $o_i$ and $m_i$ denote the observed and missing components of sample $i$. These subscripts can be used for matrices too, \textit{e.g.}, $\Sigmab_k^{o_i m_i}$ refers to the elements of the matrix $\Sigmab_k$ in the rows and columns specified by $o_i$ and $m_i$. Thus, the covariance matrix for $\xb_i$ is defined as 

\begin{equation}
	\Sigmab_{k} = 
	\begin{pmatrix} 
	\Sigmab_{k}^{o_i o_i} & \Sigmab_{k}^{o_i m_i} \\
	\Sigmab_{k}^{m_i o_i} & \Sigmab_{k}^{m_i m_i}
	\end{pmatrix}
	\label{eq:sigma_def_mm}
\end{equation}
where $\Sigmab_{k}$ has the same structure for any vector $\xb_i$, with submatrices $\Sigmab_{k}^{o_i o_i}$,  $\Sigmab_{k}^{o_i m_i}$, $\Sigmab_{k}^{m_i o_i}$ and $\Sigmab_{k}^{m_i m_i}$ changing according to the number of missing data (that depends on $i$). For brevity, we will denote $o_i = o$ and $m_i = m$ in the following, but the reader should keep in mind that these subscripts are sample-dependent.

Note that the proposed algorithm assumes data missing completely at random (MCAR) or missing at random (MAR) (see \cite{little:rubin:2002} for a detailed description of the different missing data mechanisms), which is a standard assumption for EM-based algorithms. This implies that the missing data mechanism is ignorable, \textit{i.e.}, the missingness is independent of all the values (MCAR) or independent of the missing values (MAR). In practice, the MCAR or MAR assumptions apply to a wide range of data. As a first example, the presence of clouds in remote sensing images induce missing data that may be regarded as MAR since the spatial distribution of clouds is independent of the land cover \citep{salberg2011}. As a result, any cloud detection task will not change the distribution of the observed data. A second example concerns missing pixels due to sensor failure. This setting is clearly MCAR since the missing-data mechanism is independent of the observed and missing data. In both MAR and MCAR settings, valid inferences can be obtained by ignoring the missing-data mechanism \citep{little:rubin:2002} instead of using an ad hoc procedure (data deletion, mean imputation, \textit{etc.}). Consequently, the use of the EM algorithm is justified by the assumption that the probability that a value is missing does not depend on the missing-data value itself. Finally, it would be interesting to study the case where the data is missing not at random (MNAR), such as in \cite{sportisse2021modelbased}, which is left for future work.

\subsection{The E-step}

The E-step for elliptical distributions with missing data requires to evaluate $E[\log \mathcal{L}_c(\thetab; \mathcal{X}, \mathcal{Z})) \vert \thetab^{(t)}, \mathcal{X}^o]$. Similarly to the complete data case, this can be done by focusing only on the terms $\ell_{0k}$ defined in \autoref{eq:l0}, which means that the parameters $\tau_{ik}$ can be ignored. This leads us to the following proposition.
\begin{prop}\label{prop:max_tau}
Maximizing the complete log likelihood with respect to $\thetab_k$ and $\mathcal{X}^o$ is equivalent to maximising: $$E \left[\log \mathcal{L}_0(\thetab; \mathcal{X}, \mathcal{Z}) \vert \thetab^{(t)}, \mathcal{X}^o \right] = \sum_{i=1}^{N} \sum_{k=1}^{K} E \left[z_{ik} \ell_{0k}(\xb_i ; \pi_k, \mub_k, \Sigmab_k) \vert \thetab^{(t)}, \xb_i^o \right]$$
\end{prop}
\begin{proof} See Appendix~\ref{appendix:proof_tau} for details. \end{proof}

This result is of crucial importance, since it implies that the estimation problem reduces to the estimation of a mixture of AG distributions ($\log \mathcal{L}_0(\thetab; \mathcal{X}, \mathcal{Z})$ is the log-likelihood of an AG distribution, whose pdf is provided in \autoref{eq:pdf_AG_distribution}). This leads to:

\begin{equation}
\begin{split}
    E\left[\log \mathcal{L}_0(\thetab; \mathcal{X}, \mathcal{Z}) \vert \thetab^{(t)}, \mathcal{X}^o \right] & = \sum_{i=1}^{N} \sum_{k=1}^{K} \Bigg( E\left[z_{ik} \vert \thetab^{(t)}, \xb_i^o \right] \Bigg. \\
    & \Bigg. \times E \left[\ell_{0k}(\xb_i ; \pi_k, \mub_k, \Sigmab_k) \vert z_{ik}=1, \thetab^{(t)}, \xb_i^o \right] \Bigg),
\end{split} \label{eq:missing}
\end{equation}
where the expression of $E[z_{ik} \vert \thetab^{(t)}, \xb_i^o]$ is similar to the case without missing data, except that $\hat{p}_{ik}$ is estimated using the observed data:
\begin{equation}
    \hat{p}_{ik} = \frac{\pi_k \det(\Sigmab_k^{oo})^{-1/2} \left[(\xb_i^o - \mub_k^o)^T (\Sigmab_k^{oo})^{-1}(\xb_i^o - \mub_k^o)\right]^{-d^o/2}}{\sum_{j=1}^{K} \pi_j \det(\Sigmab_j^{oo})^{-1/2} \left[(\xb_i^o - \mub_j^o)^T (\Sigmab_k^{oo})^{-1}(\xb_i^o - \mub_j^o)\right]^{-d^o/2}}, \label{eq:AngD}
\end{equation}
with $d^o$ the number of observed features for each sample. 

\subsubsection{Conditional expectations}

As can be observed in \autoref{eq:missing}, additional terms coming from $E \left[\ell_{0k}(\xb_i ; \pi_k, \mub_k, \Sigmab_k) \vert z_{ik}=1, \thetab^{(t)}, \xb_i^o \right]$ have to be computed with respect to the case without missing data. More precisely, we need to compute two new sufficient statistics, $E[\xb_i^m \vert z_{ik}=1, \xb_i^o, \thetab]$ and $\text{E}[\xb_i^m (\xb_i^m)^T\vert z_{ik}=1, \xb_i^o, \thetab]$ which are first and second order conditional expectations of the missing variables for a sample $\xb_i$, given that $\xb_i$ has been generated by the AG $k$ (this is in fact similar to the GMM case, but with different expectations, see \cite{Ghahramani_tech_report} for a detailed example in that case). These sufficient statistics can be determined easily after identifying the conditional distribution $f_{i, \thetab_k}(\xb_i^m\vert\xb_i^o)$. Using \autoref{prop:max_tau}, we have noted that the pdf $f_{i,\thetab_k}(x_i)$ reduces to an AG distribution, which is an outstanding result. Based on this, the conditional mean and covariance of $\xb_i^m \vert \xb_i^o, \thetab$ can be determined using the following proposition.

\begin{prop}\label{prop:conditional_dist}
Suppose that $\xb \sim AG_m(\mub, \Sigmab)$, with $\xb=[\xb_1^T, \xb_2^T]^T \in \mathbb{R}^d$, and $d=d_1+d_2$. Then, $\xb_2 \vert \xb_1~\sim~t_\nu(\mub_{2.1}, \Sigmab_{22.1})$, where $t_\nu(\mub_{2.1}, \Sigmab_{22.1})$ is a multivariate t-distribution with $\nu = d_1$ degrees of freedom, mean vector $\mub_{2.1} = \mub_2 + \Sigmab_{21}\Sigmab_{11}^{-1}(\xb_1 - \mub_1)$ and scale matrix $\Sigmab_{22.1}=s_{22.1}(\Sigmab_{22} - \Sigmab_{21}\Sigmab_{11}^{-1}\Sigmab_{12})$ with $s_{22.1}=\frac{1}{d_1}(\xb_1 - \mub_1)^T (\Sigmab_{11})^{-1}(\xb_1 - \mub_1) \in \mathbb{R}$.
\end{prop}
\begin{proof} See Appendix~\ref{sec:proof_conditional}.\end{proof}

\begin{remark}

\textit{This proposition is in agreement with some results obtained in the general case of elliptical distributions, denoted as $E_d(\mub, \boldsymbol{\Lambda})$ (see for instance \cite[Chapter 13]{Bilodeau_1999} for detailed proofs). More precisely, if $\xb =[\xb_1^T, \xb_2^T]^T \sim E_d(\mub, \boldsymbol{\Lambda})$, then $\xb_2 \vert \xb_1~\sim~E_{d_2}(\mub_{2.1}, \boldsymbol{\Lambda}_{22.1})$ with $\mub_{2.1} = \mub_2 + \boldsymbol{\Lambda}_{21}\boldsymbol{\Lambda}_{11}^{-1}(\xb_1 - \mub_1)$ and $\boldsymbol{\Lambda}_{22.1} = \boldsymbol{\Lambda}_{22} - \boldsymbol{\Lambda}_{21}\boldsymbol{\Lambda}_{11}^{-1}\boldsymbol{\Lambda}_{12}$. In that case, the covariance matrix of $\xb_2 \vert \xb_1$ is $\text{cov}[\xb_2 \vert \xb_1] = w(\xb_1)\boldsymbol{\Lambda}_{22.1}$, with $w$ a function depending on $\xb_1$. This paper shows that within the EM framework for elliptical distributions with missing data, the estimation of the model parameters $(\mub_k, \Sigmab_k, \pi_k)$ of any elliptical distribution reduces to consider an Angular Gaussian distribution, when assuming that the density generator is the same for each component. Further work based on this result could be interesting, but such investigations are out of the scope of this paper.}
\end{remark} 

\bigskip
The expectation of the log-likelihood can then be derived as follows:
\begin{equation}\label{complete-loglikelihood-mm}
\begin{split}
    E\left[\log \mathcal{L}_0(\thetab; \mathcal{X}, \mathcal{Z}) \vert \thetab^{(t)}, \mathcal{X}^o \right] = \sum_{i=1}^{N} \sum_{k=1}^{K} \hat{p}_{ik} & \Bigg( \log(\boldsymbol{\pi}_k) + \log(A_{i}) -\frac{1}{2}\log(\det(\Sigmab_k)) \Bigg. \\ & \Bigg. - \frac{m}{2} \log \left[\textrm{tr}((\Sigmab_k)^{-1}\Tilde{{\Sigmab}}_{ik})\right]  \Bigg. \\ & \Bigg. - \frac{m}{2} \log \left[(\Tilde{\xb}_i - \mub_k)^T \Sigmab_k^{-1}(\Tilde{\xb}_i - \mub_k)\right] \Bigg),
\end{split}
\end{equation}
where $\Tilde{{\Sigmab}}_{ik}$ and $\Tilde{\xb}_i $ are defined hereafter. \autoref{complete-loglikelihood-mm} is obtained by computing the following sufficient statistics thanks to the conditional distributions found in \autoref{prop:conditional_dist}:

\begin{align}
    \mathbb{E}[\xb_i^m \vert z_{ik}=1, \xb_i^o, \boldsymbol{\thetab}] = \mub_{ik}^{m} = & \mub_k^m + \Sigmab_k^{mo}(\Sigmab_k^{oo})^{-1}(\xb_i^o - \mub_k^o),
    \label{eq:conditional_mean}\\
    \begin{split}
    \mathbb{E}[\xb_i^m (\xb_i^m)^T\vert z_{ik}=1, \xb_i^o, \boldsymbol{\thetab}] = \Sigmab_{ik}^{mm} = & C_i^o \times (\Sigmab_k^{mm} - \Sigmab_k^{mo}(\Sigmab_k^{oo})^{-1}\Sigmab_k^{mo}),
    \end{split}
    \label{eq:conditional_cov}
\end{align}

where $\Sigmab_k^{oo}$, $\Sigmab_k^{mo}$, $\Sigmab_k^{om}$ and $\Sigmab_k^{mm}$ are properly defined in \autoref{eq:sigma_def_mm} for each sample and $C_i^o = \frac{(\xb_i^o - \mub_k^o)^T (\Sigmab_k^{oo})^{-1} (\xb_i^o - \mub_k^o)}{d^o-2}$.

As for the classical EM for GMM with missing data \citep{Ghahramani_1994, EIROLA201432}, the missing components of each $\boldsymbol{x}_i$ are replaced by the conditional mean ${\mub}_{ik}^{m}$, along with the computation of the quantity $\Tilde{\Sigmab}_{ik}$ given by
\begin{align}
	\Tilde{\xb}_{i} &= (\xb_i^o, \mub_{ik}^{m})
	\label{eq:x_hat}\\
	\Tilde{\Sigmab}_{ik} &= \begin{pmatrix} 
	\boldsymbol{0}^{oo} & \boldsymbol{0}^{om} \\
	\boldsymbol{0}^{mo} & \Sigmab_{ik}^{mm}
	\end{pmatrix}
	\label{eq:sigma_hat}
\end{align}
where $\boldsymbol{0}^{oo}, \boldsymbol{0}^{om}$ and $\boldsymbol{0}^{mo}$ are matrices of zeros of appropriate dimensions. Note that the tilde symbol is used here to highlight the terms related to the conditional expectations computed in \autoref{eq:conditional_mean} and \autoref{eq:conditional_cov}, similarly to \cite{EIROLA201432}.

\subsection{The M-step}\label{sec:M_step_missing}

Maximizing the log-likelihood leads to the following expressions for $\mub_k$, $\Sigmab_k$ and $\pi_k$ defined thanks to (classical) fixed-point equations: 

\begin{align}
\mub_k = & \sum_{i=1}^{N} \frac{w_{ik}p_{ik}\Tilde{\xb}_i }{\sum_{l=1}^N w_{lk} p_{lk}}, \\ \Sigmab_k = & m\sum_{i=1}^{N} \frac{p_{ik}}{\sum_{l=1}^N p_{lk}} \left( \frac{\Tilde{\Sigmab}_{ik} }{\textrm{tr}(\Sigmab_k^{-1}\Tilde{\Sigmab}_{ik} )} + w_{ik}(\Tilde{\xb}_i  - \mub_k)(\Tilde{\xb}_i  - \mub_k)^T \right),\\ \pi_k = & \frac{1}{N} \sum_{i=1}^N p_{ik},
\end{align}

where $w_{ik} = \frac{1}{(\Tilde{\xb}_i - \mub_k)^T \Sigmab_k^{-1}(\Tilde{\xb}_i - \mub_k)}$. \\

These expressions are intuitive, in the sense that they follow the same logic as the EM for GMM with missing data. Indeed, in the M-step, missing values are replaced by their imputed values in $\Tilde{\xb}_i$, and covariances matrices are updated using an additional term taking into account the missing values. One can also notice that those expressions are fixed-point equations as often in robust approaches (\textit{e.g.}, for $M$- or FEM estimators). 

Finally, note that the scale of $\Sigmab_k$ has no influence on the estimation of missing values (\textit{i.e.}, the scale of $\Sigmab_k$ does not change the result in \autoref{eq:conditional_mean}). This is an important result since, as explained in \autoref{sec:M_step_complete}, the scatter matrix $\Sigmab_k$ is equal to the covariance matrix up to a scale factor.

\subsection{Proposed algorithm}

The proposed generalized FEM algorithm is detailed in Algorithm~\ref{alg:fem_proposed}. As for Algorithm~\ref{alg:cap}, a more efficient way to initialize $\boldsymbol{\thetab}^{(0)}$ when compared to a random initialization is to use the K-means algorithm. In that case, a first imputation of the missing values is needed, for instance using imputations based on the mean or $k$-nearest neighbors. In the presence of missing data, the EM algorithm comes with a higher computational cost (it is true for GMM as well) mainly because one needs to evaluate $(\Sigmab_k^{oo})^{-1}$ for each sample with missing data. See \cite{delalleau2018efficient} for an interesting discussion regarding this issue. 

\begin{algorithm}[htp!]
\footnotesize
\caption{Scheme of the generalized FEM algorithm in the the incomplete data case.}\label{alg:fem_proposed}
\textbf{Input:} Data  $\{\textbf{\textit{x}}_i\}_{i=1}^n$, $K$ the number of clusters \\
\textbf{Output:} Clustering labels $\mathcal{Z}=\{z_i\}_{i=1}^n$, model parameters $\thetab = \{\pi_1, ..., \pi_k, $ $\boldsymbol{\mu}_1, ..., \boldsymbol{\mu}_k, \boldsymbol{\Sigma}_1, ..., \boldsymbol{\Sigma}_K \}$ and imputed samples $\Tilde{\xb}_i$
\begin{algorithmic}[1]
\State For each sample, identify observed and missing components $o$ and $m$;
\State Initialize $\thetab^{(0)}$;
\State $t \leftarrow 1$

\While{not convergence}

\For{$1 \leq k \leq K$:}\Comment{\textbf{E-step}}

    \Comment{Compute observed responsibilities:}
    \State $p_{ik}^{(t)} = \frac{\pi_k^{(t-1)} \det(\Sigmab_k^{oo(t-1)})^{-1/2} \left[(\xb_i^{o} - \mub_k^{o(t-1)})^T (\Sigmab_k^{oo(t-1)})^{-1}(\xb_i^{o} - \mub_k^{o(t-1)})\right]^{-d^o/2}}{\sum_{j=1}^{K} \pi_j \det(\Sigmab_j^{oo(t-1)})^{-1/2} \left[(\xb_i^o - \mub_j^{o(t-1)})^T (\Sigmab_k^{oo(t-1)})^{-1}(\xb_i^{o} - \mub_j^{o(t-1)})\right]^{-d^o/2}}$
    \medskip
    
    \Comment{Compute conditional expectations:}
    \State $\mub_{ik}^{m(t)} = \mub_k^{m(t-1)} + \Sigmab_k^{mo(t-1)}(\Sigmab_k^{oo(t-1)})^{-1}(\xb_i^o - \mub_k^{o(t-1)})$
    \medskip
    \State $\Sigmab_{ik}^{mm(t)} = \frac{(\xb_i^o - \mub_k^{o(t)})^T (\Sigmab_k^{oo(t)})^{-1} (\xb_k^o - \mub_k^{o(t)})}{d^o-2} \times (\Sigmab_k^{mm(t)} - \Sigmab_k^{mo(t)}(\Sigmab_k^{oo(t)})^{-1}\Sigmab_k^{mo(t)})$

    \State Fill in: $\Tilde{\xb}_{ik}^{(t)} \leftarrow [x_i^{o}, \mub_{ik}^{m(t)}]$ and  $\Tilde{\Sigmab}_{ik}^{(t)} \leftarrow \begin{pmatrix} 
	\boldsymbol{0}^{oo} & \boldsymbol{0}^{om} \\
	\boldsymbol{0}^{mo} & \Sigmab_{ik}^{mm(t)}
	\end{pmatrix}$
\EndFor

\For{$1 \leq k \leq K$:}\Comment{\textbf{M-step}}
\medskip
\State $\pi_{k}^{(t)} = \frac{1}{N} \sum_{i=1}^N p_{ik}^{(t)}$
\medskip
\State set $\mub_k^{'} = \mub_k^{(t-1)}$ and $\Sigmab_k^{'} = \Sigmab_k^{(t-1)}$
    \medskip
    \While{not convergence}\Comment{Fixed-point loop}
    \State $w_{ik} = \frac{1}{(\Tilde{\xb}_{ik}^{(t)} - \mub_k^{'})^T (\Sigmab_k^{'})^{-1}(\Tilde{\xb}_{ik}^{(t)} - \mub_k^{'})},$
    \medskip
    \State $\mub_k^{''} = \sum_{i=1}^{N} \frac{w_{ik}p_{ik}^{(t)} \Tilde{\xb}_{ik}^{(t)}}{\sum_{l=1}^N w_{lk}p_{lk}^{(t)}},$
    \medskip
    \State $\Sigmab_k^{''} = m\sum_{i=1}^{N} \frac{p_{ik}^{(t)}}{\sum_{l=1}^N p_{lk}^{(t)}} \left( \frac{\Tilde{\Sigmab}_{ik}^{(t)} }{\textrm{tr}((\Sigmab_k^{'})^{-1}\Tilde{\Sigmab}_{ik}^{(t)} )} + w_{ik}(\Tilde{\xb}_i^{(t)}  - \mub_k^{'})(\Tilde{\xb}_i^{(t)}  - \mub_k^{'})^T \right);$
    \medskip
    \State Update: $\mub_k^{'} = \mub_k^{''}$ and $\Sigmab_k^{'} = \Sigmab_k^{''}$
    \EndWhile
    \State Update: $\mub_k^{(t)} = \mub_k^{''}$ and $\Sigmab_k^{(t)} = \Sigmab_k^{''}$
\EndFor

\State $t \leftarrow t+1 $

\EndWhile
\State Set $z_i$ as the index $k$ that has the maximimum $p_{ik}$.
\State \textbf{Final imputation:} $\Tilde{\xb}_i^m = \sum_{k=1}^K p_{ik}^{(t)} \Tilde{\xb}_{ik}^{(t)}$
\end{algorithmic}
\end{algorithm}

\newpage
\section{Numerical results}\label{sec:results}

This section evaluates the proposed method for the reconstruction of missing data coming from various datasets used in biomedical analysis and remote sensing.

\subsection{Experimental setup}

In the following experiments, we assume that the missing data mechanism can be ignored, \textit{i.e.}, we consider the missing values to be MCAR or MAR (see discussion in \autoref{sec:method}). The imputation performance is quantitatively evaluated using the Mean Absolute Percentage Error (MAPE), which is convenient to use and interpret and can be computed with features having different magnitudes. The MAPE is defined as follows:
\begin{equation}
    \textrm{MAPE} = \frac{100}{N_{m}}\sum_{i=1}^{N_m}\frac{\vert f_{i} - \hat{f}_{i}\vert }{\vert f_i \vert},
\end{equation}
where $N_{m}$ is the number of missing features, $f_i$ is the actual value of the $i$th feature and $\hat{f}_{i}$ its estimation (also known as imputation or reconstruction). Other metrics, such as the mean absolute error (MAE) or the root mean squared error (RMSE) were also tested but are not reported here since they lead to similar conclusions (with a simpler interpretation for MAPE). Finally, note that when the features are very close to zero, the MAPE metric should be carefully used (see for instance the discussion on the abalone dataset).

The FEM algorithm is compared to five other imputation methods, namely the $k$-nearest neighbor (KNN) imputation \citep{Troyanskaya2001}, the Multiple Imputation by Chained Equations (MICE) \citep{Buuren2011}, the MissForest imputation algorithm \citep{Stekhoven_missforest_2011}, the GMM method \citep{DEMP1977} and a robust version of GMM \citep{Mouret_2021_gmm}. The robust GMM algorithm uses an outlier detection algorithm (namely, the isolation forest algorithm \citep{Liu2012}) within the EM algorithm to reduce the influence of outlier samples in the estimation of the mixture model (see the original paper \cite{Mouret_2021_gmm} for more details and derivations). For KNN, MICE and MissForest algorithms, we used the Python library \textit{scikit-learn} \citep{scikit-learn} (version 0.24.2), whereas we have implemented our own EM algorithm for GMM, robust GMM and FEM. A minimal parameter tuning was considered for the different algorithms. To that extent, KNN, MICE, MissForest algorithms were used with their default parameters: the number of neighbors was set to $5$ for the KNN algorithm, the MICE algorithm uses the IterativeImputer with BayesianRidge estimators and the MissForest uses the IterativeImputer with ExtraTree regressor. The parameters of the GMM algorithms were adjusted as in \cite{Mouret_2021_gmm} (in brief, only a small regularization of the covariances matrices is applied to avoid instabilities) and the FEM algorithm was used without any tuning (\textit{i.e.}, no regularization was used for the covariance matrices). Finally, the number of components $K$ used for the GMM and FEM depends on the datasets (it is fixed when the number of classes is known, otherwise it is estimated using the Bayesian Information Criterion, see results on the abalone dataset for details).  

In a first experiment, two synthetic datasets are considered to evaluate a change in the underlying data distribution (\textit{i.e.}, Gaussian and non-Gaussian distributions). In a second step, imputation tasks are conducted on different real world datasets. We mostly used benchmark datasets coming from the University of California at Irvine (UCI) database\footnote{\url{https://archive.ics.uci.edu}}. To that extent, the results presented here can be easily reproduced. The name, number of attributes and number of samples of each dataset are summarized in \autoref{table:datasets}. For all these datasets, 50 Monte Carlo (MC) simulations were conducted by varying the percentage of missing data and the percentage of outliers added to the dataset. The outliers are generated using uniform distributions on intervals defined by the minimum and maximum of each feature.

\begin{table}[htp]
    
    \caption{Numbers of features and samples of the datasets used to evaluate the different imputation techniques. UCI is added in parenthesis for the datasets coming from the UCI database.}
    \centering
    \begin{tabular}{llll}
         Dataset & Features & Samples & Reference \\ \hline
         Synthetic (AR(1) time series) & 10 & 2000 & - \\
         Mice protein expression (UCI) & 82 & 1080 & \cite{Higuera_2015} \\
         Abalone (UCI) & 8 & 4176 & \cite{nash1994population} \\ 
         Statlog - Landsat Satellite (UCI) & 36 & 4435 (train) & - \\
         Rapeseed crops - Sentinel satellites & 106 & 2218 & \cite{Mouret_2021}\\
    \end{tabular}
    \label{table:datasets}
\end{table}

\subsection{Synthetic data}

The proposed imputation algorithm is first tested on synthetic data. The conducted experiments can be summarized as follows: 1) generation of synthetic data with $N=2000$ (number of samples) and $m=10$ (number of features), 2) scaling of the generated data (this allows us to use the MAPE metric without problems), 3) random generation of missing values (the missing elements of the feature matrix are chosen randomly), 4) imputation (and metric computation). The datasets are generated according to a mixture of $K=3$ distributions (all classes are equally likely represented). The generation of these datasets is summarized in \autoref{table:synthetic_parameters} (which provides the distributions used for the datasets) and \autoref{table:synthetic_generation} (which presents the generation of the synthetic samples). Note that after having generated the synthetic samples, the dataset is scaled so that the minimum value of the whole dataset is equal to 1 and the $98$th\footnote{This values was chosen so that most of the data (except outliers) are scaled in the range $[1, 100]$.} percentile of the whole dataset is equal to $100$ (this allow us to avoid problems with the MAPE metrics when values are close to zero).

\begin{table}[htp!]
    \caption{Summary of the parameters used to generate the synthetic samples, where $\mathcal{U}$ is the uniform distribution.}
    \centering
    \begin{tabular}{llll}
         Parameters \\ \hline
         $\mu_{kf} \sim \mathcal{U}[0, 1]$, $\forall f$ with f feature $\#$f \\ \hline
         $\phi_k \sim \mathcal{U}[0.1,0.9]$ \\
         $\sigma_k^2 \sim \mathcal{U}[0.0005,0.005]]$ \\
         
         $\Sigmab_k = \frac{\sigma_k^2}{1 - \phi_k^2}\begin{pmatrix} 
                        1 & \phi_k & \phi_k^2 & ... & \phi_k^{m-1} \\
                        \phi_k & 1 & \phi_k & ... & \phi_k^{m-1} \\
                        \phi_k^2 & \phi_k & 1 & ... & \phi_k^{m-1} \\
                        ... & ... & ... & ... & ... \\
                        \phi_k^{m-1} & ... & \phi_k^2 & \phi_k & 1 \\
                        \end{pmatrix} $
    \end{tabular}
    \label{table:synthetic_parameters}
\end{table}

\begin{table}[htp!]
    \caption{Distributions used to generate the two synthetic datasets, which make use of the parameters provided in \autoref{table:synthetic_parameters}, where $t_{5}$ is the multivariate $t$-distribution with $5$ degrees of freedom.}
    \centering
    \begin{tabular}{lll}
         Name & Generating distribution & Classes \\ \hline
         Gaussian dataset & $\mathcal{N}(\mub_k, \Sigmab_k)$ & 3 \\
         Student dataset & $t_{5}(\mub_k, \Sigmab_k)$ & 3
    \end{tabular}
    \label{table:synthetic_generation}
\end{table}

\begin{figure}[htp!]
    \centering
    \subfloat[]{\includegraphics[width=0.5\textwidth]{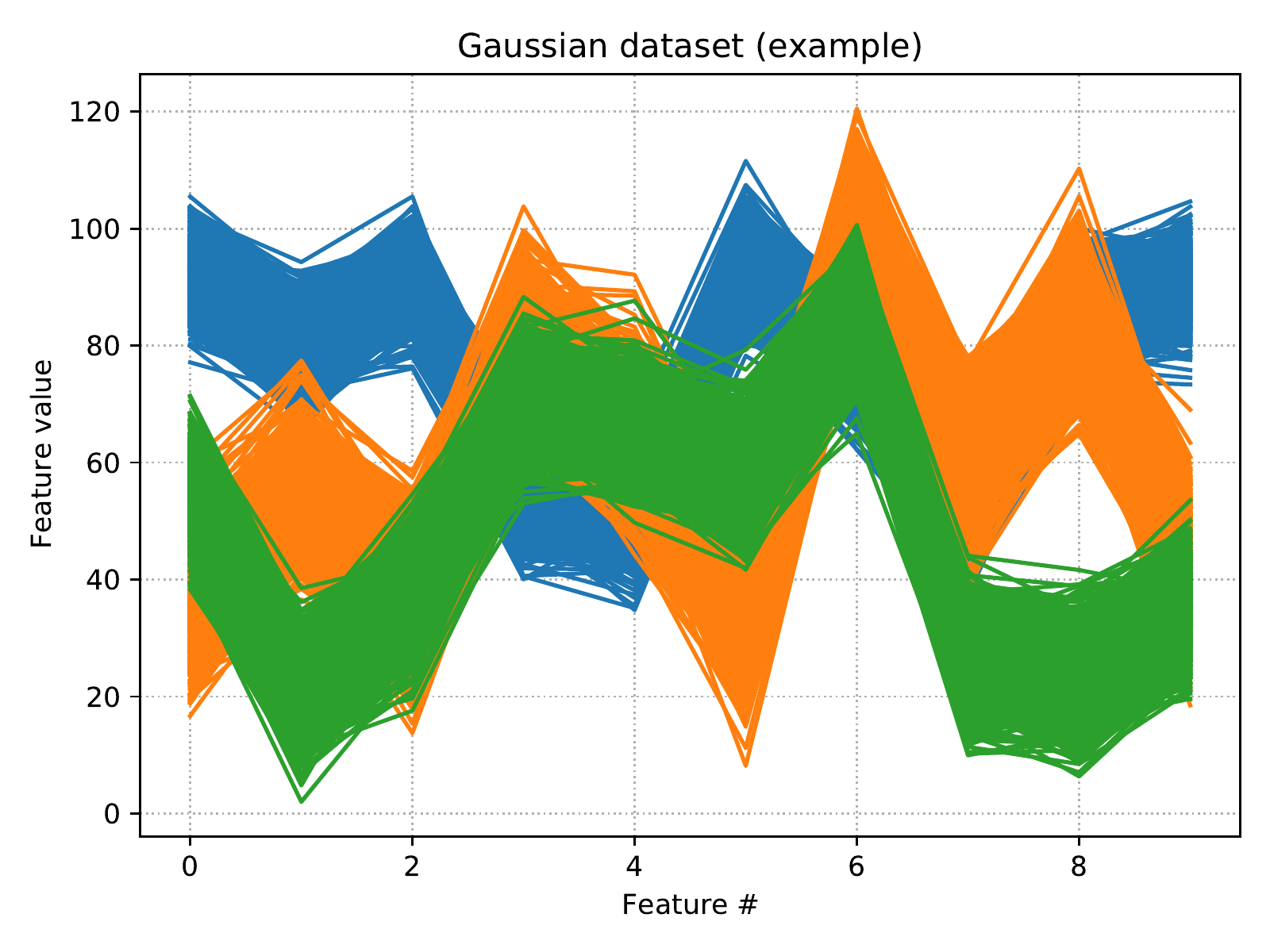}}
    \subfloat[]{\includegraphics[width=0.5\textwidth]{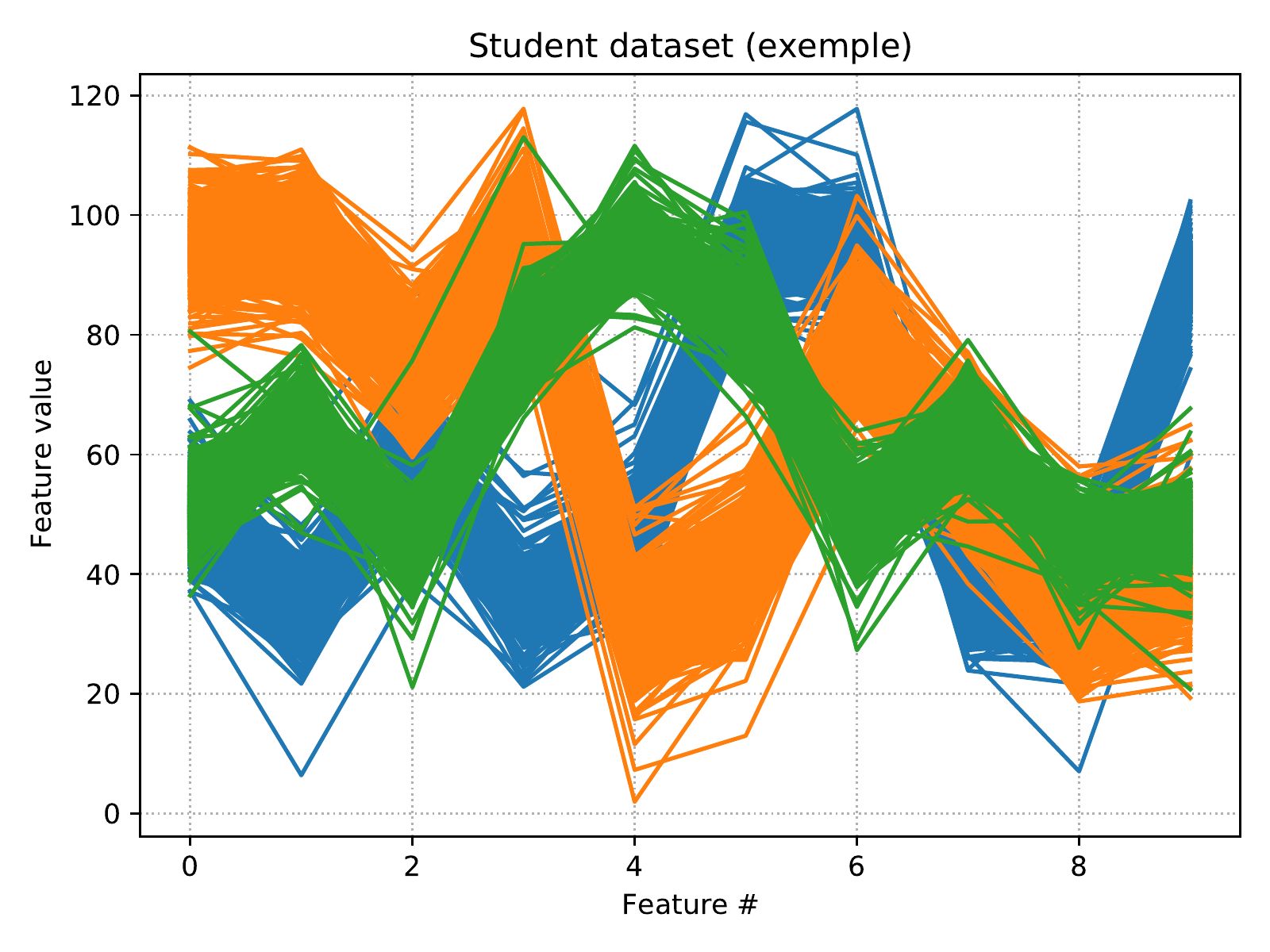}}
    \caption{Synthetic datasets generated according to \autoref{table:synthetic_generation}: a) multivariate Normal distribution, b) multivariate $t$-distribution. Each color corresponds to a different class.}
    \label{fig:examples_synthetic_datasets}
\end{figure}

The various synthetic datasets can be viewed as AR(1) time series, which are grouped into 3 different clusters. Two representative examples generated according to the Gaussian and Student's $t$-distributions are displayed in \autoref{fig:examples_synthetic_datasets}. One can observe that the Student dataset logically contains more samples with extreme values. Because of these extreme values, the range of the MAPE is different for these two types of datasets (since the data have been scaled).

\begin{figure}[htp!]
    \centering
    \subfloat[]{\includegraphics[width=0.5\textwidth]{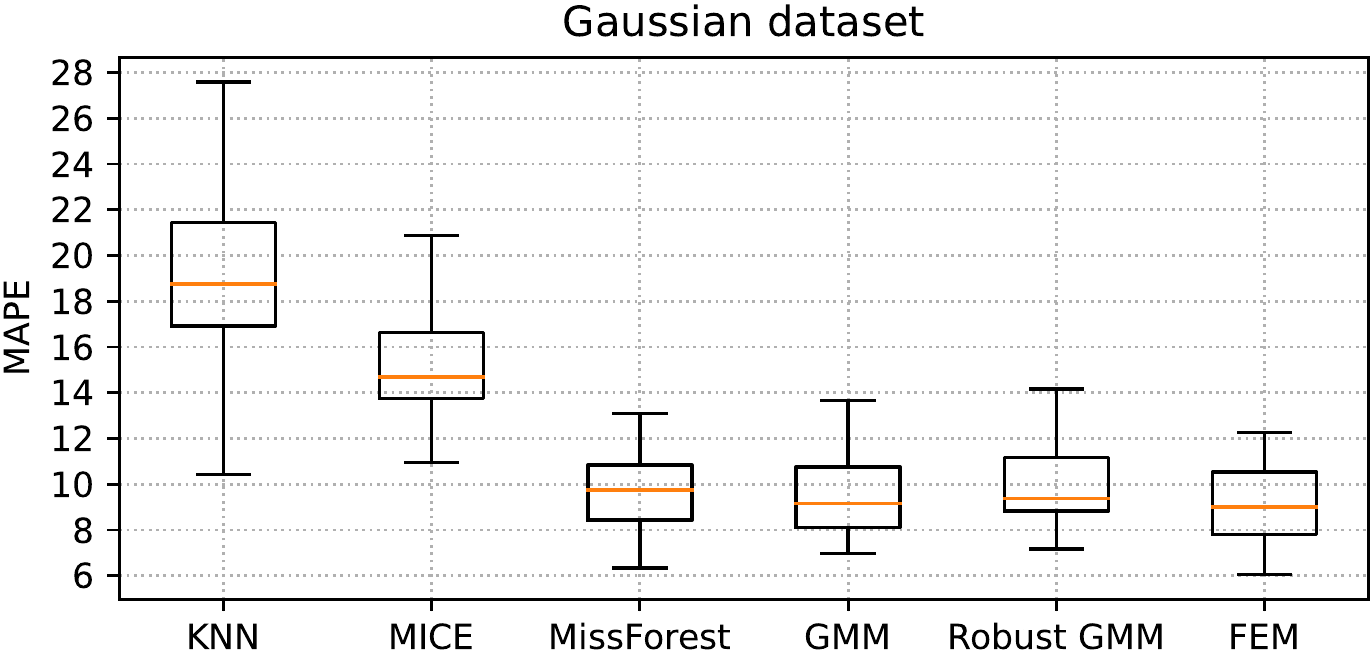}}
    \subfloat[]{\includegraphics[width=0.5\textwidth]{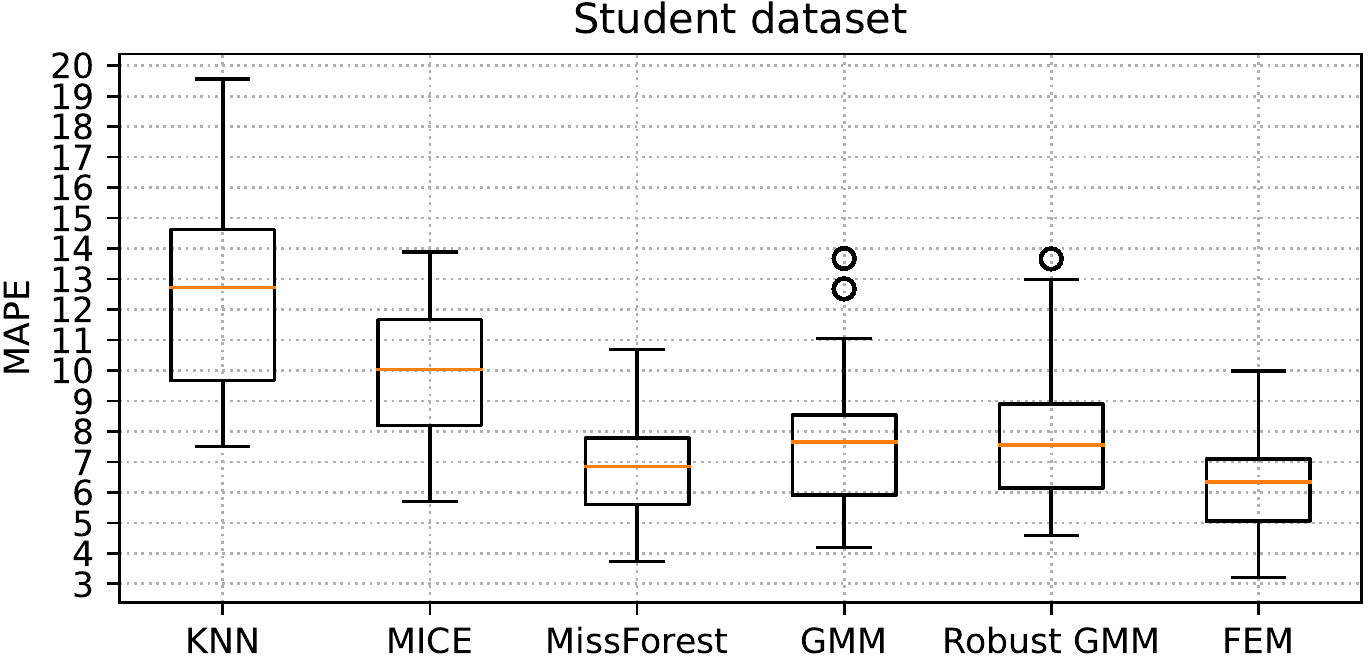}}
    \caption{MAPE obtained after 50 MC simulations a) Gaussian and b) Student synthetic datasets (generated randomly for each simulation). The percentage of missing features is 50\% for this example.}
    \label{fig:MAPE_synthetic_data}
\end{figure}

Imputation results obtained for the two types of datasets are provided in \autoref{fig:MAPE_synthetic_data} for a percentage of missing values equal to 50\%, without outliers (more experiments conducted with different percentages of missing data are provided for the real world datasets). For the Gaussian datasets, the GMM and FEM methods provide the best results, with a slight advantage for the FEM approach. This illustrates the versatility of the FEM algorithm, which is competitive against the GMM imputations, even when it is used for a dataset adapted to Gaussian methods. When considering the Student datasets, FEM still provides the best results and outperforms all the other tested approaches. In that case, the GMM approaches can provide good imputations depending on the dataset generation, but can also lead to poor results. Moreover, the robust GMM does not improve the imputation results for this type of data (this is probably due to a non-optimal tuning of the algorithm parameters used for robust estimation). In both scenarios, the MissForest provides competitive results, but with a higher MAPE than the one obtained using the FEM approach. Finally, for both types of datasets, MICE and KNN approaches are not competitive, with a MAPE that can be almost twice higher than with FEM imputations.

\subsection{Real-world experiments}
We now proceed with experiments conducted on real-world data sets.

\subsubsection{Mice protein expression dataset}

The mice protein dataset contains expression levels of various protein or protein modification measures in the cerebral cortex of $8$ classes of control and down syndrome mice. This type of data is subject to the missing data problem, which is interesting in the context of this study. In particular, the dataset considered here already contains a small percentage of missing values (around 1\%). During the experiments, missing values were simulated completely at random. The number of clusters was fixed to $8$ for the GMM and FEM algorithms.

\begin{figure}[H]
    \centering
    \textbf{Mice protein expression dataset}
    \includegraphics[width=1\textwidth]{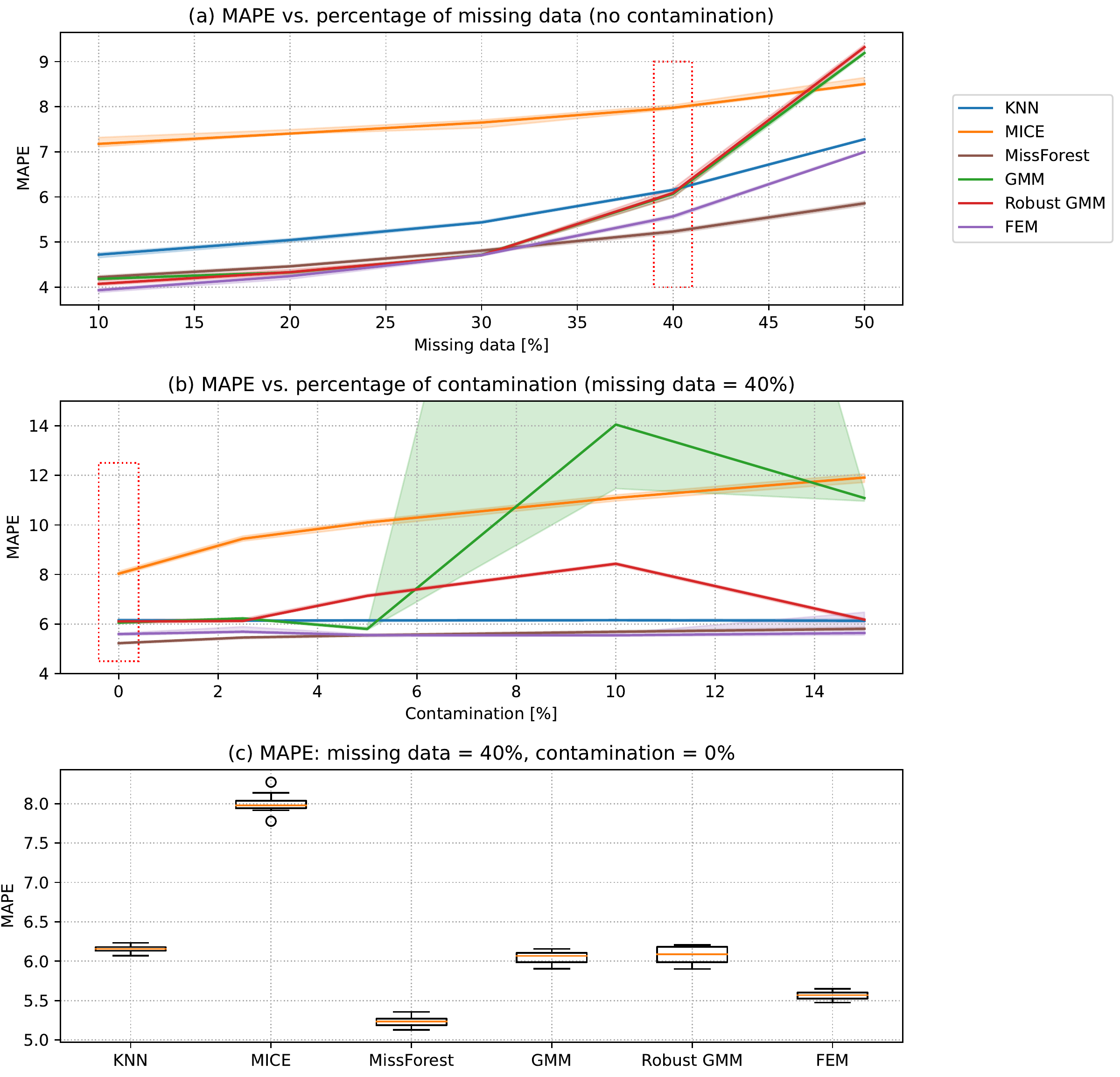}
    \caption{MAPE obtained for the mice protein expression dataset by varying (a) the percentage of missing data (MCAR) and (b) the percentage of outliers in the dataset. The results are obtained after 50 MC simulations (the plain line corresponds to the median and the shaded area is filled between the first and third quartiles). Figure (c) shows a boxplot of the MC runs when the quantity of missing data is set to 40\% and there is no outlier, which corresponds to the red rectangle displayed in (a) and (b).}
    \label{fig:panel_mice_protein}
\end{figure}

The imputation results obtained with the different methods are summarized in \autoref{fig:panel_mice_protein}. More precisely, \autoref{fig:panel_mice_protein}(a) evaluates the influence of the percentage of missing data whereas \autoref{fig:panel_mice_protein}(b) studies the influence of adding outliers in the dataset. It can be observed that, overall, the FEM and MissForest algorithm outperform all the other tested algorithms. More precisely, for percentages of missing values lower than 35\%, the FEM imputation is optimal whereas for higher percentages of missing data the MissForest algorithm provides the best imputation. The GMM approaches are competitive only when the percentage of missing data is lower than 30\%, with an important degradation for higher levels of missing data (in that case the FEM algorithm is much more competitive than the other EM-based algorithms). Moreover, GMM are particularly sensitive to the presence of outliers (for a better visualization, the MAPE obtained for GMM is not fully displayed in \autoref{fig:panel_mice_protein}(b) since some values are close to 80\%). For this dataset, the KNN and MICE algorithms perform poorly (it is especially true for the MICE algorithm). Finally, \autoref{fig:panel_mice_protein}(c) shows a boxplot representation of the MAPE obtained for different MC runs (missing data is set to 40\% without outliers). It can be observed that the MissForest algorithm always provides a better MAPE in that configuration and that the MAPE obtained using the FEM algorithm is very close.

\subsubsection{Abalone dataset}

The abalone dataset consists of various physical measurements (length, diameter, etc.) made on abalone. Note that the feature ``sex'' is a categorical feature taking the value M, F or I (Infant), which was not considered in this experiment. A simple way to handle this feature would be to convert it into integers. However, this conversion would provide inconsistent results for the MAPE or other reconstruction metrics. Note also that the feature ``rings'', which is an integer directly related to the age of the abalone, was kept in the dataset. Because some features can be very close to zero, a scaling of each feature in the range [1, 100] was made in order to use the MAPE metrics in a relevant way. This scaling make senses since the aim of these experiments is to compare the different imputation methods. Finally, the optimal number of clusters to be chosen for EM approaches was estimated for each simulation using the Bayesian Information Criterion (BIC), as recommended for instance in \cite{BOUVEYRON_review}. It is defined as $\text{BIC} = -2\log(L) + p\log(N)$ (the lower the better), where $L$ corresponds to the likelihood of a given model, $p$ is the number of parameters and $N$ is the number of samples used to fit the GMM parameters. When estimating the number of classes, we observed that the FEM algorithm tends to use efficiently a higher number of components when compared to GMM approaches, which on the contrary tend to be very unstable when the number of components is too high. Further investigations on that topic could be interesting but are out of the scope of this paper. 

\begin{figure}[htp!]
    \centering
    \textbf{Abalone dataset}\par\medskip
    
    \includegraphics[width=1\textwidth]{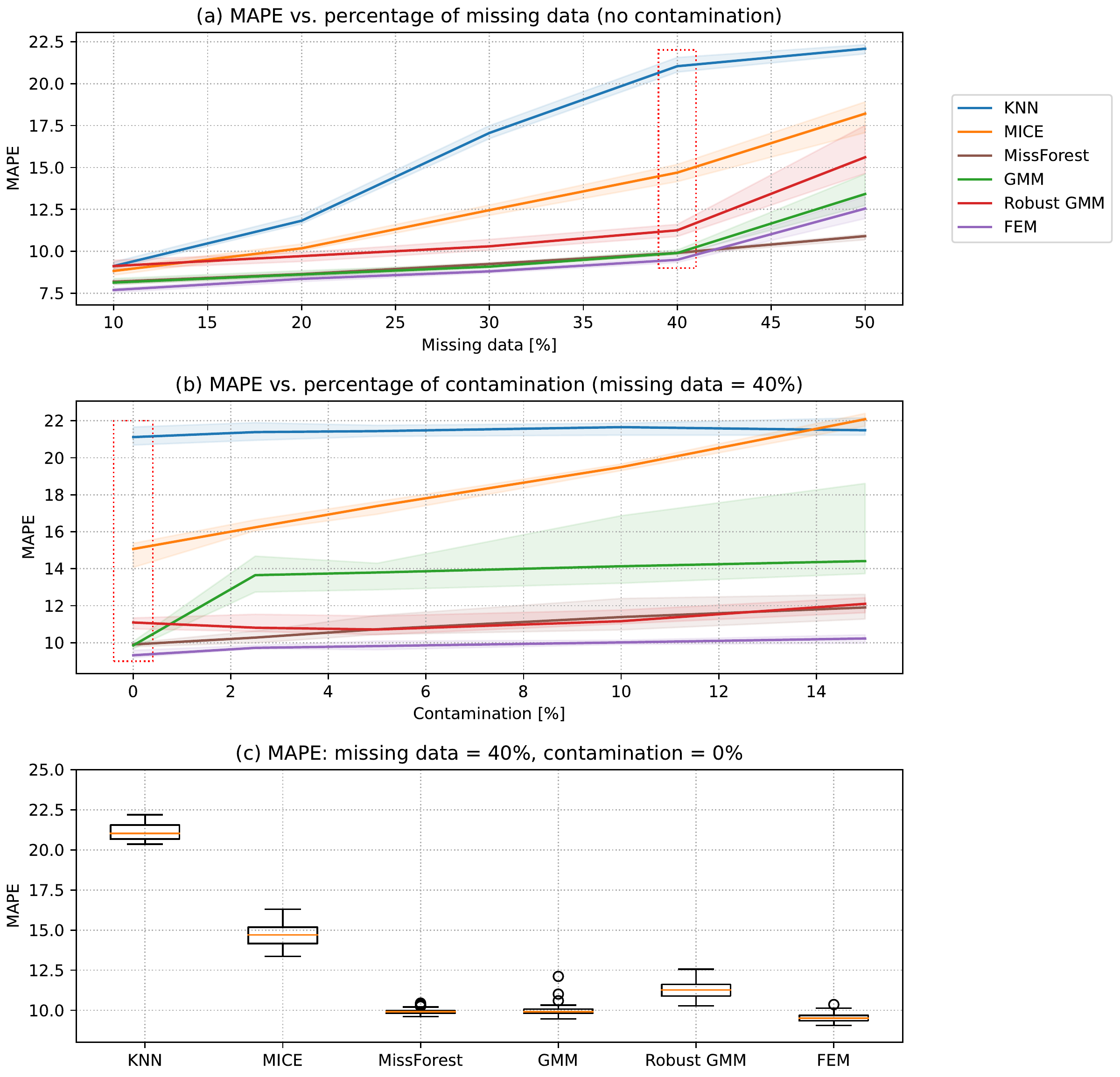}
    \caption{MAPE obtained on the abalone dataset by varying the (a) percentage of missing data (MCAR) and (b) the percentage of outliers in the dataset. The results have been obtained after averaging 50 MCs simulations (the plain line corresponds to the median and the shaded area is filled between the first and third quartiles). Figure (c) displays a boxplot of the MC runs when the quantity of missing data is 40\% without outliers added to the dataset, which corresponds to the red rectangles displayed in (a) and (b).}
    \label{fig:panel_abalone}
\end{figure}

The results obtained on the abalone dataset are summarized in \autoref{fig:panel_abalone}. In brief, most of the conclusions obtained for the mice dataset can be transposed to the abalone dataset. More precisely, the imputations obtained with the FEM algorithm have overall a lower MAPE when compared to the other algorithms. In particular, the FEM algorithm performs well when the percentage of missing data is high or when the data is contaminated by outliers, with a very low dispersion in its results. As an exception, the MissForest provides better results when the percentage of missing data is equal to 50\%. This algorithm is, however, more sensitive than the FEM algorithm to the presence of outliers. In absence of outliers, GMM imputations are close to those obtained with the FEM algorithm, but are always sub-optimal (the MAPE is consistently higher of around 0.5\%). Finally, the MICE algorithm outperforms the KNN imputation method for this dataset. This example confirms that MICE and KNN algorithms are very sensitive to the considered dataset.

\subsubsection{Statlog - Landsat satellite data}

\begin{figure}[htp!]
    \centering
    \textbf{Landsat dataset}\par\medskip
    \includegraphics[width=1\textwidth]{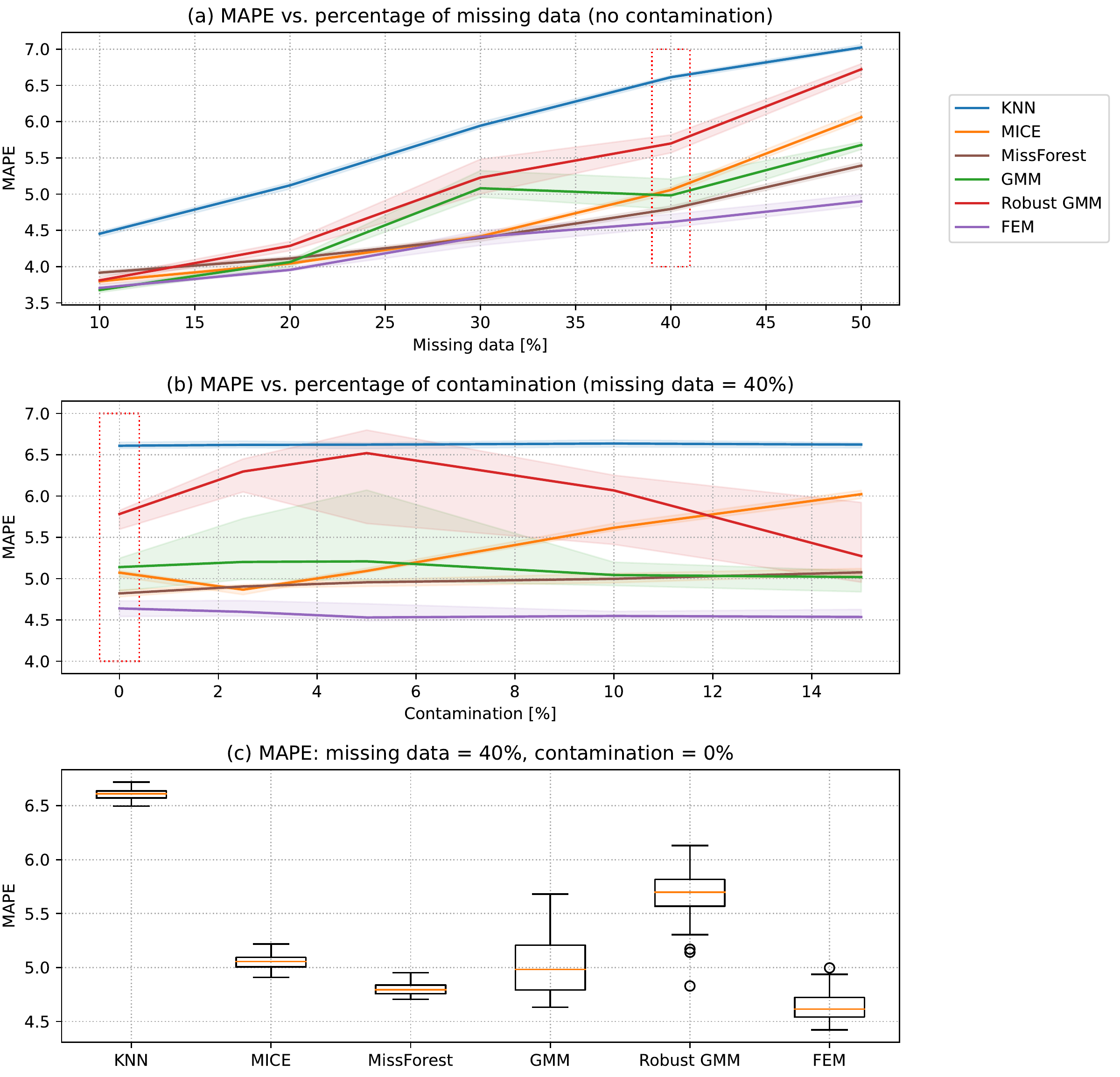}
    \caption{MAPE obtained for the Landsat dataset by varying (a) the percentage of missing data (MCAR) and (b) the percentage of outliers in the dataset. The results have been obtained after averaging the metrics of 50 MCs simulations (the plain line corresponds to the median and the shaded area is filled between the first and third quartiles). Figure (c) shows a boxplot of the MC runs when the quantity of missing data is 40\% without outliers in the dataset, which corresponds to the red rectangles plotted in (a) and (b).}
    \label{fig:panel_landsat}
\end{figure}

This classic database contains multispectral pixel values ($4$ spectral bands) acquired in a $3\times3$ neighborhood region using the Landsat satellite. Each sample is characterized by a total of $3\times3\times4 = 36$ features, which belong to a land cover category (6 categories in total, \textit{e.g.}, red soil, cotton crop, etc.). For this dataset, the missing data was simulated by removing all the values of some pixels (\textit{i.e.}, all the spectral bands are missing), which could for instance correspond to a sensor failure. Experimental results obtained on this dataset are summarized in \autoref{fig:panel_landsat}. Except for the KNN algorithm, all methods provide decent results when the percentage of missing data is lower than 20\% (with a slight and consistent advantage for the FEM algorithm). However, for higher percentages or in the presence of outliers, results obtained with the FEM algorithm are significantly better (with a low dispersion). This confirms the results obtained on the datasets previously tested. As a last remark, we would like to clarify that the relatively poor results obtained with the robust GMM approach might be explained by a non-optimal tuning of the outlier detection mechanisms, whereas the FEM algorithm is not impacted at all by outliers. This illustrates the advantage of the FEM approach, which does not need additional tuning to take into account the presence of outlier samples.

\newpage
\subsubsection{Rapeseed crop monitoring - Sentinel satellite data}

This section considers remote sensing time series computed for the monitoring of $2218$ rapeseed parcels. The time series are obtained using Sentinel-1 (S1) and Sentinel-2 (S2) satellites, which provide synthetic aperture radar (SAR) and multispectral images, respectively. The dataset is subject to missing data, especially because clouds affect multispectral images, which is a known issue in remote sensing \citep{Shen2015}. The time series to be imputed are the median and interquartile range (IQR) of statistics (computed at the parcel-level) of the Normalized difference vegetation index (NDVI), which is a popular agronomic indicator used in remote sensing for agricultural applications. The feature matrix used in these experiments also contains features coming from S1 images, which are not subject to missing data. More precisely, these features are the median (computed at the parcel-level) of the VV and VH backscattering coefficients (see \citep{Mouret_2021} for more details regarding the construction of this feature matrix). To summarize, each rapeseed parcel is characterized by $13$ values (each value corresponds to a specific time instant in the growing season of interest) of median NDVI, 13 values of IQR NDVI, 40 values of median VV backscattering and 40 values of VH backscattering (\textit{i.e.}, a total of $106$ features).

For this dataset, the missing values were not completely added at random to have more realistic experiments. Indeed, missing data occurs on cloudy days and only affect the multispectral features (here, the NDVI statistics computed at the parcel-level). More precisely, two parameters control the missing data mechanism: the percentage of multispectral images affected by missing data (\textit{i.e.}, the number of cloudy multispectral images), and the percentage of crop parcels with missing data (\textit{i.e.}, generally, only a part of the image is covered by clouds). For each multispectral image with missing data, we fixed the percentage of affected parcels to 50\%. The number of mixture components used in GMM and FEM is unknown and was fixed using BIC as for the abalone dataset.

\begin{figure}[H]
    \centering
    \textbf{Rapeseed crop monitoring - Sentinel satellite data}\par\medskip
    \includegraphics[width=1\textwidth]{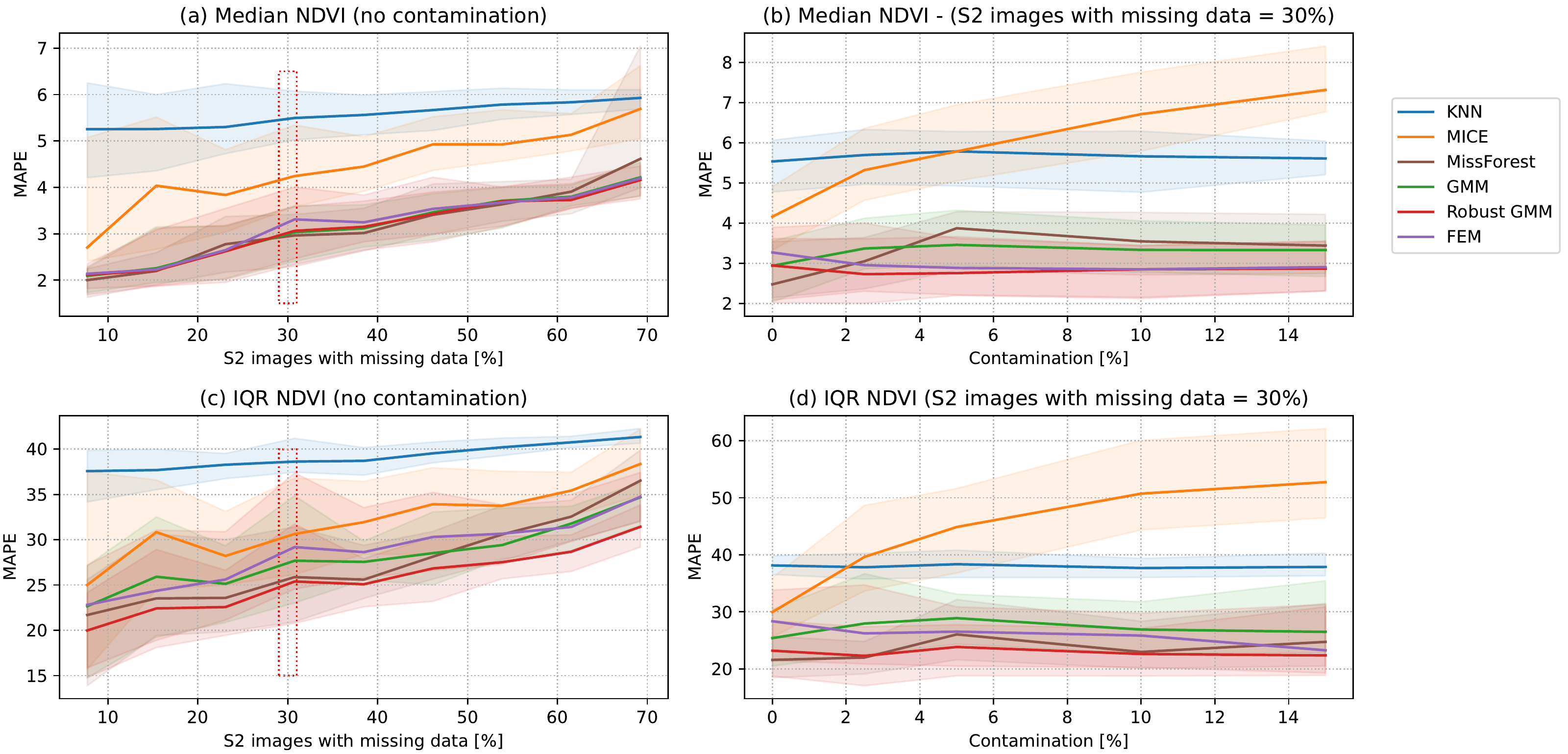}
    \caption{MAPE obtained for the rapeseed dataset by varying (a, c) the percentage of S2 images affected by missing data and (b, d) the percentage of outliers in the dataset (red boxes in (a,c) correspond to the percentage of missing S2 images used for (b,d)). When an S2 image has missing data, 50\% of the parcels have their corresponding features missing. The results are obtained after averaging the metrics of 50 MC simulations (the plain line corresponds to the median and the shaded area is filled between the first and third quartiles). Figures (a,b) are obtained using the median NDVI of the parcels whereas (c,d) corresponds to the IQR NDVI of the parcels.}
    \label{fig:panel_rapeseed}
\end{figure}

Results computed using 50 MC simulations are summarized in \autoref{fig:panel_rapeseed}, when looking separately at the median NDVI (a,b) and the IQR NDVI (c,d). Some general observations are first provided. The imputation results are more scattered than with the other datasets (one explanation is that some periods of the growing season are more difficult to reconstruct, see \cite{Mouret_2021_gmm} for more details).  However, very good reconstructions of the median NDVI are possible, even with a high percentage of S2 images with missing data, in part due to the use of additional S1 data. Regarding the IQR NDVI, the high values of the MAPE can be explained by 1) the fact that IQR NDVI values are close to zero (\textit{i.e.}, a small imputation error implies a large MAPE) and 2) the fact that IQR NDVI can change abruptly through time and is less correlated to S1 data.

When looking specifically at each algorithm, it appears that, overall, the robust GMM algorithm is the best suited for this dataset, confirming previous results found in \cite{Mouret_2021_gmm}. It can be observed that, as for the other datasets, methods based on the EM algorithm outperform the KNN and MICE methods. The MissForest algorithm performs again well even if it is impacted by outliers, confirming previous results. Finally, the FEM algorithm provides results that are very close to the GMM imputations. This is interesting since the FEM algorithm was used without any deep tuning of its parameters, whereas the robust GMM was mainly designed and tested for this task with an accurate parameter tuning.

\subsection{Influence of different types of outliers}

\begin{figure}[htp]
    \centering
    \textbf{Abalone dataset}
    
    \subfloat[]{\includegraphics[width=0.8\textwidth]{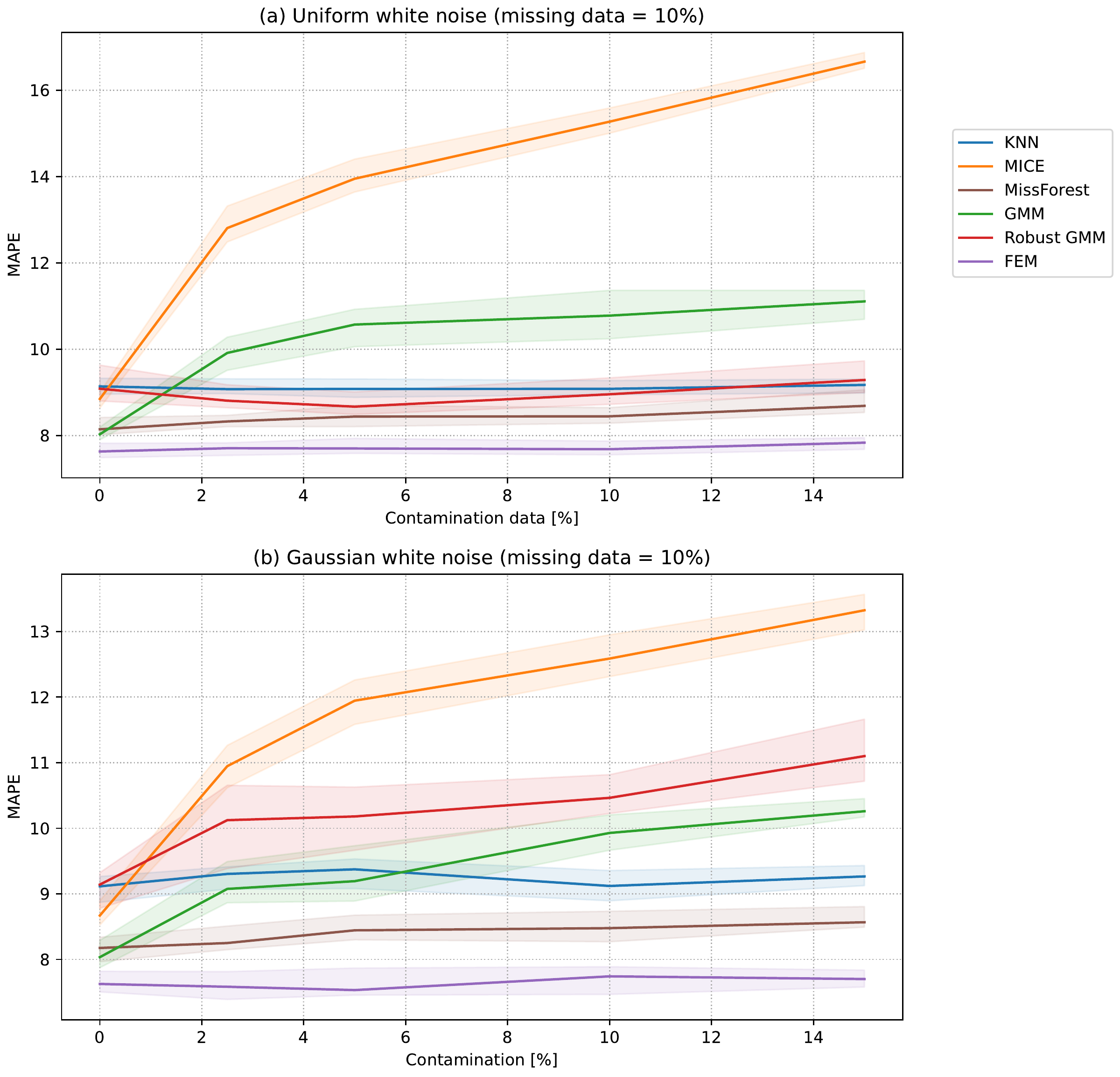}}
    \caption{MAPE obtained for the Abalone dataset by varying the percentage of outliers in the dataset, with a percentage of missing data equal to 10\%. Outliers are generated according to (a) a uniform distribution taking values between the minimum and maximum of the features and (b) a Gaussian noise (\textit{i.e.}, $z_i \sim \mathcal{N}(\boldsymbol{\mu}, \text{diag}\{\boldsymbol{\sigma}\})$, with $\boldsymbol{\mu}$ a vector whose element are the mean of the features and $\text{diag}\{\boldsymbol{\sigma}\}$ a diagonal matrix whose element are the variances of the features. The results have been obtained after averaging the metrics of $50$ Monte Carlo simulations (the plain line corresponds to the median and the shaded area is filled between the first and third quartiles).}
    \label{fig:new_expe}
\end{figure}

Two additional experiments are conducted using the Abalone dataset to 1) evaluate the impact of outliers with a low percentage of missing data (10\%) and 2) evaluate the impact of another outlier generation mechanism. More precisely, we used a mechanism similar to the one proposed in \cite{HIPPERTFERRER2022108460} by adding outliers generated as Gaussian noise (\textit{i.e.}, $\boldsymbol{z}_i \sim \mathcal{N}(\boldsymbol{\mu}, \text{diag}\{\boldsymbol{\sigma})\}$, with $\boldsymbol{\mu}$ a vector whose elements are the mean of each feature and $\text{diag}\{\boldsymbol{\sigma}\}$ a diagonal matrix whose elements are the variances of each feature. The obtained results are summarized in \autoref{fig:new_expe}.

Overall, two main conclusions can be drawn. First, even with a low percentage of missing data, imputation results obtained with the FEM algorithm are very competitive. Note that even with a small amount of missing data, the impact of outliers can be important depending on some algorithms (\textit{e.g.}, classical GMM). Secondly, changing the outlier generation mechanism has an impact on the imputation results. In particular, the robust GMM algorithm is more impacted by Gaussian white noise, while it is not the case with uniform white noise. Overall, these results confirm that the FEM algorithm is almost not impacted by outliers when changing the percentage of missing data and provide competitive results when compared to the state-of-the-art.

\newpage
\section{Conclusion}\label{sec:conclusion}

This paper proposed to extend the flexible EM (FEM) algorithm of \cite{roizman2020flexible, roizman2021_eusipco} to handle missing data. The algorithm is flexible in the sense that it is 1) robust to outliers and 2) adapted to any mixture of elliptical distributions (\textit{i.e.}, the data distribution is not necessarily Gaussian). As a consequence, the FEM algorithm can be used for a wide range of datasets, unlike the classical EM for Gaussian mixture models, which is impacted by noise and non-Gaussian distributions. The main theoretical contribution of this paper is to derive, in the presence of missing data, an EM algorithm which assumes that the data has been generated from a mixture of (unknown) elliptical distributions having the same density generator. As in the complete-data case, the FEM algorithm derived in the presence of missing data is intuitive and can be used with little parameter tuning. 

The main focus of this paper is the imputation of missing data. Imputation results obtained using the FEM algorithm were compared with 5 other benchmark algorithms, based on KNN, MICE, MissForst, GMM and robust GMM. From all the experiments presented in this study, two main conclusions can be drawn. First, it was observed that the FEM algorithm was competitive for all considered datasets, generally outperforming all other tested methods when the percentage of missing data is high or when outliers are contaminating the dataset. To that extent, the experimental results confirm the theoretical robust properties of the algorithm and illustrate the flexibility of the proposed algorithm when compared to the other tested methods, which may fail depending on the considered dataset.

Using the FEM algorithm for outlier detection and classification tasks (potentially with missing values) is an interesting prospect. Other perspectives are related to the regularization of the scatter matrix, which can be complicated to estimate with high dimensional data \citep{BOUVEYRON_review}. Various regularization strategies used for GMM could be investigated for FEM models, such as the approaches proposed in \cite{BOUVEYRON_article} to regularize the eigenvalues of the covariance matrices or the $l-1$ constrained graphical lasso algorithm \citep{friedman_sparse_2008}, which has been extended to GMM with missing data in \cite{ruan_2011}.

\section*{Declarations}

\begin{itemize}
\item Funding: This document is the result of a research project funded by TerraNIS SAS and ANRT (convention CIFRE no. 2018/1349)
\item Conflict of interest/Competing interests: The authors have no competing interest to declare that is relevant to the content of this article.
\item Ethics approval: not applicable
\item Consent to participate: not applicable
\item Consent for publication: not applicable
\item Availability of data and materials: We mostly used benchmark datasets coming from the University of California at Irvine (UCI) database (\url{https://archive.ics.uci.edu}).
\item Code availability: the implementation in Python of the algorithm is available at the repository \url{https://github.com/fmouret/flexible_em_imputation}.
\item Authors' contributions: All authors contributed to the study conception and design. The first draft of the manuscript was written by Florian Mouret and all authors commented on previous versions of the manuscript. All authors read and approved the final manuscript.
\end{itemize}

\begin{appendices}
\section{Proofs}

\subsection{Independence of $E[\ell_{ik}(\xb_i; \pi_k, \mub_k, \Sigmab_k, \tau_{ik}) \vert z_{ik}, \thetab^{(t)}, \xb_i^o]$ from the model parameters $(\pi_k, \mub_k, \Sigmab_k$)}\label{appendix:proof_tau}

This appendix shows the independence of $E[\ell_{ik}(\xb_i; \pi_k, \mub_k, \Sigmab_k, \tau_{ik}) \vert z_{ik}, \thetab^{(t)}, \xb_i^o]$ from the model parameters $(\pi_k, \mub_k, \Sigmab_k$). From \eqref{eq:Lc_comple_data_bis}, the expectation of the complete log-likelihood knowing the model parameters and the observed variables can be written as follows:
\begin{equation}\label{eq:E_Lc}
\begin{split}
    E\left[\log \mathcal{L}_c(\thetab; \mathcal{X}, \mathcal{Z})) \vert \thetab^{(t)}, \mathcal{X}^o\right] = \sum_{i=1}^{N} \sum_{k=1}^{K} \begin{aligned}[t] & E\Big[z_{ik}\ell_{0k}(\xb_i; \pi_k, \mub_k, \Sigmab_k) + z_{ik}\ell_{ik}(\xb_i ; \pi_k, \mub_k, \Sigmab_k, \tau_{ik}) \vert \thetab^{(t)}, \xb_i^o \Big],
    \end{aligned}
    \end{split}
\end{equation}where, similarly to the complete data case, we define $\ell_{ik}(\xb_i ; \pi_k, \mub_k, \Sigmab_k, \tau_{ik}) = \log(s_{ik}^{d/2}g_{i}(s_{ik}))$ with $s_{ik} = \cfrac{(\xb_i-\boldsymbol{\mu}_k)^{T} \boldsymbol{\Sigma}_k^{-1}(\xb_i-\boldsymbol{\mu}_k)}{\tau_{ik}}$. Using the fact that $z_{ik}$ is a binary indicator, \eqref{eq:E_Lc} can be  decomposed as follows:
\begin{equation} \label{eq:E_Lc_bis}
\footnotesize
\begin{split}
    E\left[\log \mathcal{L}_c(\thetab; \mathcal{X}, \mathcal{Z})) \vert \thetab^{(t)}, \mathcal{X}^o\right] = & \sum_{i=1}^{N} \sum_{k=1}^{K} E\left[z_{ik}\vert \thetab^{(t)}, \xb_i^o \right] E\left[\ell_{0k}(\xb_i; \pi_k, \mub_k, \Sigmab_k)\vert z_{ik}=1,\thetab^{(t)}, \xb_i^o \right] \\ & + \sum_{i=1}^{N} \sum_{k=1}^{K} E\left[z_{ik}\vert \thetab^{(t)}, \xb_i^o \right]E\left[\ell_{ik}(\xb_i ; \pi_k, \mub_k, \Sigmab_k, \tau_{ik}) \vert z_{ik}=1,\thetab^{(t)}, \xb_i^o \right].
    \end{split}
\end{equation}
Since the first term of this expression does not depend on $\tau_{ik}$, the maximization of \eqref{eq:E_Lc_bis} w.r.t. $\tau_{ik}$ reduces to maximize the following function:
\begin{equation} \label{eq:function}
\begin{split}
    E\left[\ell_{ik}(\xb_i ; \pi_k, \mub_k, \Sigmab_k, \tau_{ik}) \vert z_{ik}=1,\thetab^{(t)}, \xb_i^o \right] & = E\left[\log\left(s_{ik}^{d/2}\,g_i(s_{ik})\right) \vert z_{ik}=1, \boldsymbol{\theta}_k, \xb_i^o\right].
\end{split}
\end{equation}

\noindent Consider the one dimensional function $f$ defined by $f(t) = t^{d/2} g(t), t \in \mathbb{R}$. The maximum of this function is denoted as
$$
\displaystyle t^*=\arg\sup_t\{f(t)\},
$$
which is a constant independent of $\pi_k, \boldsymbol{\mu}_k$ and $\boldsymbol{\Sigma}_k$. Moreover, the supremum of the function $f$ is well-defined and is denoted as $\sup \{ f(t^*) \}$. Using the fact that $\log$ is an increasing function, one has
$$
\log \left(s_{ik}^{d/2}\,g_i(s_{ik} \right) \le \log[\sup(f(t^*))],
$$
which implies
$$
E\left[\ell_{ik}(\xb_i ; \pi_k, \mub_k, \Sigmab_k, \tau_{ik}) \vert z_{ik}=1,\thetab^{(t)}, \xb_i^o \right]  \le  \log[\sup(f(t^*))].
$$
Moreover, define $$\tau^{*}_{ik} = \cfrac{(\xb_i-\boldsymbol{\mu}_k)^{T} \boldsymbol{\Sigma}_k^{-1}(\xb_i-\boldsymbol{\mu}_k)}{t^*}.$$
Replacing $\tau_{ik}$ by $\tau^{*}_{ik}$ in \eqref{eq:function} leads to
\begin{equation}\label{eq:ineq}
 E\left[\ell_{ik}(\xb_i ; \pi_k, \mub_k, \Sigmab_k, \tau^{*}_{ik}) \vert z_{ik}=1,\thetab^{(t)}, \xb_i^o \right] =\log[\sup(f(t^*))].
\end{equation}
This shows that the conditional expectation is maximized for $\tau_{ik} = \tau^{*}_{ik}$ and that this maximum does not depend on the model parameters $\pi_k, \boldsymbol{\mu}_k, \boldsymbol{\Sigma}_k$ and the missing data. Thus, maximizing $E\left[\log \mathcal{L}_c(\thetab; \mathcal{X}, \mathcal{Z})) \vert \thetab^{(t)}, \mathcal{X}^o\right] $ w.r.t. $\thetab$ and $\mathcal{X}^o$ is equivalent to maximizing the AG part of the log-likelihood, i.e., $\sum_{i=1}^{N} \sum_{k=1}^{K} E \left[z_{ik} \ell_{0k}(\xb_i ; \pi_k, \mub_k, \Sigmab_k) \vert \thetab^{(t)}, \xb_i^o \right]$, which concludes the proof.

\subsection{Conditional distribution of an Angular Gaussian distribution}\label{sec:proof_conditional}

This appendix derives the conditional density $f_{i, \thetab_k} (\xb_2\vert \xb_1)$. For brevity we denote $\xb_i = \xb$, $\Sigmab_k = \Sigmab$ and $\mub_k = \mub$ in the following. By definition of a conditional pdf, we have
\begin{equation}
    f_{\thetab} (\xb_2 \vert \xb_1) = \frac{f_{\thetab} (\xb_1, \xb_2)}{f_{\thetab} (\xb_1)},
\end{equation}
\textit{i.e.},
\begin{equation}\label{eq:conditional_dist}
    f_{\thetab} (\xb_2\vert \xb_1) \propto \frac{\det(\Sigmab)^{-1/2} \left[(\xb - \mub)^T \Sigmab^{-1}(\xb - \mub)\right]^{-d/2}}{\det(\Sigmab_{11})^{-1/2} \left[(\xb_1 - \mub_1)^T (\Sigmab_{11})^{-1}(\xb_1 - \mub_1)\right]^{-d_1/2}}
\end{equation}
where $d_1$ is the number of features in $\xb_1$.

The determinant of the matrix $\Sigmab$ can be decomposed as
\begin{equation}
    \det(\Sigmab^{-1/2}) = [\det(\Sigmab_{22} - \Sigmab_{21}\Sigmab_{11}^{-1}\Sigmab_{12})\det(\Sigmab_{11})]^{-1/2}.
\end{equation}
Moreover, by using standard manipulations on matrices, we obtain
\begin{equation}\label{eq:manip}
    \begin{split}
        \left[(\xb - \mub)\Sigmab^{-1}(\xb - \mub) \right]^{-d/2} = & \left[(\xb_1 - \mub_1)^T (\Sigmab_{11})^{-1} (\xb_1 - \mub_1) \right. \\ & + \left. \left(\xb_2 - \mub_{2.1}\right)^T \Sigmab_{22.1}^{-1}\left(\xb_2 - \mub_{2.1}\right) \right]^{-d/2},
    \end{split}
\end{equation}
with $\mub_{2.1} = \mub_2 - \Sigmab_{21}\Sigmab_{11}^{-1}(\xb_1 - \mub_1)$ and $\Sigmab_{2.1} = \Sigmab_{22} - \Sigmab_{21}\Sigmab_{11}^{-1}\Sigmab_{12}$. Thus plugin \autoref{eq:manip} into \autoref{eq:conditional_dist} gives:

\footnotesize
\begin{align}
    f_{\thetab} (\xb_2\vert \xb_1) & \propto \frac{\det(\Sigmab_{22.1})^{-1/2} \left[(\xb_1 - \mub_1)^T (\Sigmab_{11})^{-1} (\xb_1 - \mub_1) + \left(\xb_2 - \mub_{2.1}\right)^T \Sigmab_{22.1}^{-1}\left(\xb_2 - \mub_{2.1}\right) \right]^{-d/2}}{\left[(\xb_1 - \mub_1)^T (\Sigmab_{11})^{-1}(\xb_1 - \mub_1)\right]^{-d_1/2}} \\
   & \propto  \frac{\det(\Sigmab_{22.1})^{-1/2} \left[(\xb_1 - \mub_1)^T (\Sigmab_{11})^{-1} (\xb_1 - \mub_1)\right]^{-d/2}  \left[1 + \frac{\left(\xb_2 - \mub_{2.1}\right)^T \Sigmab_{22.1}^{-1}\left(\xb_2 - \mub_{2.1}\right)}{(\xb_1 - \mub_1)^T (\Sigmab_{11})^{-1} (\xb_1 - \mub_1)} \right]^{-d/2}}{\left[(\xb_1 - \mub_1)^T (\Sigmab_{11})^{-1}(\xb_1 - \mub_1)\right]^{-d_1/2}} \\
 & \propto  \frac{\det(\Sigmab_{22.1})^{-1/2} \left[1 + \frac{\left(\xb_2 - \mub_{2.1}\right)^T \Sigmab_{22.1}^{-1}\left(\xb_2 - \mub_{2.1}\right)}{(\xb_1 - \mub_1)^T (\Sigmab_{11})^{-1} (\xb_1 - \mub_1)} \right]^{-d/2}}{\left[(\xb_1 - \mub_1)^T (\Sigmab_{11})^{-1}(\xb_1 - \mub_1)\right]^{\frac{d-d_1}{2}}}. \label{eq:cond_1}
\end{align}

Looking at the numerator of \autoref{eq:cond_1}, a multivariate $t$- distribution can be identified since
\begin{equation}
    f_{\text{student}}(\xb_2) \propto  \left[1 + \frac{1}{\nu}(\boldsymbol{x_2}-\mub_{\text{student}})^T\Sigmab_{\text{student}}^{-1}(\boldsymbol{x_2}-\mub_{\text{student}})\right]^{-(\nu + d_2)/2},
\end{equation}
where $\Sigmab_{\text{student}}$ is the scale matrix of $\xb_2$. By identification, the following results are obtained:

$$
\nu = d_1
$$
$$\mub_{\text{student}} = \mub_{2.1} = \mub_2 + \Sigmab_{21}\Sigmab_{11}^{-1}(\xb_1 - \mub_1)
$$
$$
\Sigmab_{\text{student}}^{-1} = \frac{d_1}{(\xb_1 - \mub_1)^T (\Sigmab_{11})^{-1} (\xb_1 - \mub_1)} \times \Sigmab_{22.1}^{-1}
$$
   $$
   \Sigmab_{\text{student}} = \frac{(\xb_1 - \mub_1)^T (\Sigmab_{11})^{-1}(\xb_1 - \mub_1)}{d_1} \times \Sigmab_{22.1}
   $$

\end{appendices}

\bibliographystyle{cas-model2-names}
\bibliography{references}

\begin{thebibliography}{44}
\expandafter\ifx\csname natexlab\endcsname\relax\def\natexlab#1{#1}\fi
\providecommand{\url}[1]{\texttt{#1}}
\providecommand{\href}[2]{#2}
\providecommand{\path}[1]{#1}
\providecommand{\DOIprefix}{doi:}
\providecommand{\ArXivprefix}{arXiv:}
\providecommand{\URLprefix}{URL: }
\providecommand{\Pubmedprefix}{pmid:}
\providecommand{\doi}[1]{\href{http://dx.doi.org/#1}{\path{#1}}}
\providecommand{\Pubmed}[1]{\href{pmid:#1}{\path{#1}}}
\providecommand{\bibinfo}[2]{#2}
\ifx\xfnm\relax \def\xfnm[#1]{\unskip,\space#1}\fi
\bibitem[{Anderson(1957)}]{Anderson_1957}
\bibinfo{author}{Anderson, T.W.}, \bibinfo{year}{1957}.
\newblock \bibinfo{title}{Maximum likelihood estimates for a multivariate
  normal distribution when some observations are missing}.
\newblock \bibinfo{journal}{J. Am. Stat. Assoc.} \bibinfo{volume}{52},
  \bibinfo{pages}{200--203}.
\newblock \DOIprefix\doi{10.1080/01621459.1957.10501379}.
\bibitem[{{Bilodeau} and {Brenner}(1999)}]{Bilodeau_1999}
\bibinfo{author}{{Bilodeau}, M.}, \bibinfo{author}{{Brenner}, D.},
  \bibinfo{year}{1999}.
\newblock \bibinfo{title}{{Theory of multivariate statistics}}.
\newblock \bibinfo{publisher}{Springer, New York}.
\newblock \DOIprefix\doi{10.1007/b97615}.
\bibitem[{Bouveyron and Brunet-Saumard(2014)}]{BOUVEYRON_review}
\bibinfo{author}{Bouveyron, C.}, \bibinfo{author}{Brunet-Saumard, C.},
  \bibinfo{year}{2014}.
\newblock \bibinfo{title}{Model-based clustering of high-dimensional data: A
  review}.
\newblock \bibinfo{journal}{Comput. Stat. Data Anal.} \bibinfo{volume}{71},
  \bibinfo{pages}{52--78}.
\newblock \URLprefix
  \url{https://www.sciencedirect.com/science/article/pii/S0167947312004422},
  \DOIprefix\doi{https://doi.org/10.1016/j.csda.2012.12.008}.
\bibitem[{Bouveyron et~al.(2007)Bouveyron, Girard and
  Schmid}]{BOUVEYRON_article}
\bibinfo{author}{Bouveyron, C.}, \bibinfo{author}{Girard, S.},
  \bibinfo{author}{Schmid, C.}, \bibinfo{year}{2007}.
\newblock \bibinfo{title}{High-dimensional data clustering}.
\newblock \bibinfo{journal}{Comput. Stat. Data Anal.} \bibinfo{volume}{52},
  \bibinfo{pages}{502--519}.
\newblock \URLprefix
  \url{https://www.sciencedirect.com/science/article/pii/S0167947307000692},
  \DOIprefix\doi{https://doi.org/10.1016/j.csda.2007.02.009}.
\bibitem[{Browne and McNicholas(2015)}]{Browne_2015}
\bibinfo{author}{Browne, R.P.}, \bibinfo{author}{McNicholas, P.D.},
  \bibinfo{year}{2015}.
\newblock \bibinfo{title}{A mixture of generalized hyperbolic distributions}.
\newblock \bibinfo{journal}{Can. J. Stat.} \bibinfo{volume}{43},
  \bibinfo{pages}{176--198}.
\newblock \DOIprefix\doi{https://doi.org/10.1002/cjs.11246}.
\bibitem[{van Buuren(2018)}]{van2018flexible}
\bibinfo{author}{van Buuren, S.}, \bibinfo{year}{2018}.
\newblock \bibinfo{title}{Flexible imputation of missing data}.
\newblock \bibinfo{publisher}{CRC press}.
\newblock \DOIprefix\doi{https://doi.org/10.1201/9780429492259}.
\bibitem[{van Buuren and Groothuis-Oudshoorn(2011)}]{Buuren2011}
\bibinfo{author}{van Buuren, S.}, \bibinfo{author}{Groothuis-Oudshoorn, K.},
  \bibinfo{year}{2011}.
\newblock \bibinfo{title}{{MICE}: Multivariate imputation by chained equations
  in {R}}.
\newblock \bibinfo{journal}{J. Stat. Softw.} \bibinfo{volume}{45},
  \bibinfo{pages}{1--67}.
\newblock \URLprefix \url{https://www.jstatsoft.org/v045/i03},
  \DOIprefix\doi{10.18637/jss.v045.i03}.
\bibitem[{Campbell(1984)}]{Campbell_1984}
\bibinfo{author}{Campbell, N.A.}, \bibinfo{year}{1984}.
\newblock \bibinfo{title}{Mixture models and atypical values}.
\newblock \bibinfo{journal}{Math. Geol.} \bibinfo{volume}{16},
  \bibinfo{pages}{465--477}.
\newblock \DOIprefix\doi{10.1007/BF01886327}.
\bibitem[{Cismondi et~al.(2013)Cismondi, Fialho, Vieira, Reti, Sousa and
  Finkelstein}]{CISMONDI201363}
\bibinfo{author}{Cismondi, F.}, \bibinfo{author}{Fialho, A.S.},
  \bibinfo{author}{Vieira, S.M.}, \bibinfo{author}{Reti, S.R.},
  \bibinfo{author}{Sousa, J.M.}, \bibinfo{author}{Finkelstein, S.N.},
  \bibinfo{year}{2013}.
\newblock \bibinfo{title}{Missing data in medical databases: Impute, delete or
  classify?}
\newblock \bibinfo{journal}{Artif. Intell. Med.} \bibinfo{volume}{58},
  \bibinfo{pages}{63--72}.
\newblock \DOIprefix\doi{https://doi.org/10.1016/j.artmed.2013.01.003}.
\bibitem[{Conte et~al.(2002)Conte, De~Maio and Ricci}]{Conte_2002}
\bibinfo{author}{Conte, E.}, \bibinfo{author}{De~Maio, A.},
  \bibinfo{author}{Ricci, G.}, \bibinfo{year}{2002}.
\newblock \bibinfo{title}{Covariance matrix estimation for adaptive {CFAR}
  detection in compound-{G}aussian clutter}.
\newblock \bibinfo{journal}{IEEE Trans. Aerosp. Electron. Syst.}
  \bibinfo{volume}{38}, \bibinfo{pages}{415--426}.
\newblock \DOIprefix\doi{10.1109/TAES.2002.1008976}.
\bibitem[{Delalleau et~al.(2018)Delalleau, Courville and
  Bengio}]{delalleau2018efficient}
\bibinfo{author}{Delalleau, O.}, \bibinfo{author}{Courville, A.},
  \bibinfo{author}{Bengio, Y.}, \bibinfo{year}{2018}.
\newblock \bibinfo{title}{Efficient {EM} training of {G}aussian mixtures with
  missing data}.
\newblock \href{http://arxiv.org/abs/https://arxiv.org/abs/1209.0521}{\tt
  arXiv:https://arxiv.org/abs/1209.0521}.
\bibitem[{Dempster et~al.(1977)Dempster, Laird and Rubin}]{DEMP1977}
\bibinfo{author}{Dempster, A.P.}, \bibinfo{author}{Laird, N.M.},
  \bibinfo{author}{Rubin, D.B.}, \bibinfo{year}{1977}.
\newblock \bibinfo{title}{Maximum likelihood from incomplete data via the {EM}
  algorithm}.
\newblock \bibinfo{journal}{J. R. Stat. Soc.} \bibinfo{volume}{39}.
\bibitem[{Eirola et~al.(2014)Eirola, Lendasse, Vandewalle and
  Biernacki}]{EIROLA201432}
\bibinfo{author}{Eirola, E.}, \bibinfo{author}{Lendasse, A.},
  \bibinfo{author}{Vandewalle, V.}, \bibinfo{author}{Biernacki, C.},
  \bibinfo{year}{2014}.
\newblock \bibinfo{title}{Mixture of {G}aussians for distance estimation with
  missing data}.
\newblock \bibinfo{journal}{Neurocomputing} \bibinfo{volume}{131},
  \bibinfo{pages}{32--42}.
\newblock \DOIprefix\doi{https://doi.org/10.1016/j.neucom.2013.07.050}.
\bibitem[{Farhangfar et~al.(2007)Farhangfar, Kurgan and
  Pedrycz}]{Farhangfar_2007}
\bibinfo{author}{Farhangfar, A.}, \bibinfo{author}{Kurgan, L.A.},
  \bibinfo{author}{Pedrycz, W.}, \bibinfo{year}{2007}.
\newblock \bibinfo{title}{A novel framework for imputation of missing values in
  databases}.
\newblock \bibinfo{journal}{IEEE Transactions on Systems, Man, and Cybernetics
  - Part A: Systems and Humans} \bibinfo{volume}{37},
  \bibinfo{pages}{692--709}.
\newblock \DOIprefix\doi{10.1109/TSMCA.2007.902631}.
\bibitem[{Fraley and Raftery(2002)}]{Fraley_2002}
\bibinfo{author}{Fraley, C.}, \bibinfo{author}{Raftery, A.E.},
  \bibinfo{year}{2002}.
\newblock \bibinfo{title}{Model-based clustering, discriminant analysis, and
  density estimation}.
\newblock \bibinfo{journal}{J. Am. Stat. Assoc.} \bibinfo{volume}{97},
  \bibinfo{pages}{611--631}.
\newblock \URLprefix \url{http://www.jstor.org/stable/3085676}.
\bibitem[{Friedman et~al.(2008)Friedman, Hastie and
  Tibshirani}]{friedman_sparse_2008}
\bibinfo{author}{Friedman, J.}, \bibinfo{author}{Hastie, T.},
  \bibinfo{author}{Tibshirani, R.}, \bibinfo{year}{2008}.
\newblock \bibinfo{title}{Sparse inverse covariance estimation with the
  graphical lasso}.
\newblock \bibinfo{journal}{Biostatistics} \bibinfo{volume}{9},
  \bibinfo{pages}{432--441}.
\newblock \DOIprefix\doi{10.1093/biostatistics/kxm045}.
\bibitem[{Ghahramani and Jordan(1994a)}]{Ghahramani_1994}
\bibinfo{author}{Ghahramani, Z.}, \bibinfo{author}{Jordan, M.},
  \bibinfo{year}{1994}a.
\newblock \bibinfo{title}{Supervised learning from incomplete data via an {EM}
  approach}, in: \bibinfo{booktitle}{Advances in Neural Information Processing
  Systems}, \bibinfo{publisher}{Morgan-Kaufmann}. pp.
  \bibinfo{pages}{120--127}.
\newblock \URLprefix
  \url{https://proceedings.neurips.cc/paper/1993/file/f2201f5191c4e92cc5af043eebfd0946-Paper.pdf}.
\bibitem[{Ghahramani and Jordan(1994b)}]{Ghahramani_tech_report}
\bibinfo{author}{Ghahramani, Z.}, \bibinfo{author}{Jordan, M.I.},
  \bibinfo{year}{1994}b.
\newblock \bibinfo{title}{Learning from Incomplete Data}.
\newblock \bibinfo{type}{Technical Report}. Massachusetts Institute of
  Technology.
\newblock \URLprefix \url{http://mlg.eng.cam.ac.uk/zoubin/papers/review.pdf}.
\bibitem[{Higuera et~al.(2015)Higuera, Gardiner and Cios}]{Higuera_2015}
\bibinfo{author}{Higuera, C.}, \bibinfo{author}{Gardiner, K.J.},
  \bibinfo{author}{Cios, K.J.}, \bibinfo{year}{2015}.
\newblock \bibinfo{title}{Self-organizing feature maps identify proteins
  critical to learning in a mouse model of {D}own syndrome}.
\newblock \bibinfo{journal}{PLOS ONE} \bibinfo{volume}{10},
  \bibinfo{pages}{1--28}.
\newblock \DOIprefix\doi{10.1371/journal.pone.0129126}.
\bibitem[{Hippert-Ferrer et~al.(2022)Hippert-Ferrer, {El Korso}, Breloy and
  Ginolhac}]{HIPPERTFERRER2022108460}
\bibinfo{author}{Hippert-Ferrer, A.}, \bibinfo{author}{{El Korso}, M.},
  \bibinfo{author}{Breloy, A.}, \bibinfo{author}{Ginolhac, G.},
  \bibinfo{year}{2022}.
\newblock \bibinfo{title}{Robust low-rank covariance matrix estimation with a
  general pattern of missing values}.
\newblock \bibinfo{journal}{Signal Processing} \bibinfo{volume}{195},
  \bibinfo{pages}{108460}.
\newblock \DOIprefix\doi{https://doi.org/10.1016/j.sigpro.2022.108460}.
\bibitem[{Kelker(1970)}]{Kelker_1970}
\bibinfo{author}{Kelker, D.}, \bibinfo{year}{1970}.
\newblock \bibinfo{title}{Distribution theory of spherical distributions and a
  location-scale parameter generalization}.
\newblock \bibinfo{journal}{Sankhya A} \bibinfo{volume}{32},
  \bibinfo{pages}{419--430}.
\bibitem[{Lin and Tsai(2020)}]{Lin_2020}
\bibinfo{author}{Lin, W.C.}, \bibinfo{author}{Tsai, C.F.},
  \bibinfo{year}{2020}.
\newblock \bibinfo{title}{Missing value imputation: a review and analysis of
  the literature (2006-2017)}.
\newblock \bibinfo{journal}{Artif. Intell. Rev.} \bibinfo{volume}{53},
  \bibinfo{pages}{1487--1509}.
\newblock \DOIprefix\doi{10.1007/s10462-019-09709-4}.
\bibitem[{Little and Rubin(2002)}]{little:rubin:2002}
\bibinfo{author}{Little, R.J.}, \bibinfo{author}{Rubin, D.B.},
  \bibinfo{year}{2002}.
\newblock \bibinfo{title}{{Statistical analysis with missing data}}.
\newblock \bibinfo{edition}{2nd} ed., \bibinfo{publisher}{John Wiley \& Sons,
  Inc. Hoboken, NJ, USA}.
\bibitem[{Liu et~al.(2012)Liu, Ting and Zhou}]{Liu2012}
\bibinfo{author}{Liu, F.T.}, \bibinfo{author}{Ting, K.M.},
  \bibinfo{author}{Zhou, Z.H.}, \bibinfo{year}{2012}.
\newblock \bibinfo{title}{Isolation-based anomaly detection}.
\newblock \bibinfo{journal}{ACM Trans. Knowl. Discov. Data}
  \bibinfo{volume}{6}.
\newblock \DOIprefix\doi{10.1145/2133360.2133363}.
\bibitem[{Mirza et~al.(2019)Mirza, Wang, Wang, Choi, Chung and
  Ping}]{Mirza_2019}
\bibinfo{author}{Mirza, B.}, \bibinfo{author}{Wang, W.}, \bibinfo{author}{Wang,
  J.}, \bibinfo{author}{Choi, H.}, \bibinfo{author}{Chung, N.C.},
  \bibinfo{author}{Ping, P.}, \bibinfo{year}{2019}.
\newblock \bibinfo{title}{Machine learning and integrative analysis of
  biomedical big data}.
\newblock \bibinfo{journal}{Genes} \bibinfo{volume}{10}.
\newblock \DOIprefix\doi{10.3390/genes10020087}.
\bibitem[{Moran et~al.(1997)Moran, Inoue and Barnes}]{MORAN1997319}
\bibinfo{author}{Moran, M.}, \bibinfo{author}{Inoue, Y.},
  \bibinfo{author}{Barnes, E.}, \bibinfo{year}{1997}.
\newblock \bibinfo{title}{Opportunities and limitations for image-based remote
  sensing in precision crop management}.
\newblock \bibinfo{journal}{Remote Sens. Environ.} \bibinfo{volume}{61},
  \bibinfo{pages}{319--346}.
\newblock \DOIprefix\doi{https://doi.org/10.1016/S0034-4257(97)00045-X}.
\bibitem[{Mouret et~al.(2021)Mouret, Albughdadi, Duthoit, Kouamé, Rieu and
  Tourneret}]{Mouret_2021}
\bibinfo{author}{Mouret, F.}, \bibinfo{author}{Albughdadi, M.},
  \bibinfo{author}{Duthoit, S.}, \bibinfo{author}{Kouamé, D.},
  \bibinfo{author}{Rieu, G.}, \bibinfo{author}{Tourneret, J.Y.},
  \bibinfo{year}{2021}.
\newblock \bibinfo{title}{Outlier detection at the parcel-level in wheat and
  rapeseed crops using multispectral and {SAR} time series}.
\newblock \bibinfo{journal}{Remote Sens.} \bibinfo{volume}{13},
  \bibinfo{pages}{956}.
\newblock \URLprefix \url{http://dx.doi.org/10.3390/rs13050956},
  \DOIprefix\doi{10.3390/rs13050956}.
\bibitem[{Mouret et~al.(2022)Mouret, Albughdadi, Duthoit, Kouamé, Rieu and
  Tourneret}]{Mouret_2021_gmm}
\bibinfo{author}{Mouret, F.}, \bibinfo{author}{Albughdadi, M.},
  \bibinfo{author}{Duthoit, S.}, \bibinfo{author}{Kouamé, D.},
  \bibinfo{author}{Rieu, G.}, \bibinfo{author}{Tourneret, J.Y.},
  \bibinfo{year}{2022}.
\newblock \bibinfo{title}{Reconstruction of {S}entinel-2 derived time series
  using robust {G}aussian mixture models. {A}pplication to the detection of
  anomalous crop development}.
\newblock \bibinfo{journal}{Comput. Electron. Agric.} \bibinfo{volume}{198},
  \bibinfo{pages}{106983}.
\newblock \DOIprefix\doi{https://doi.org/10.1016/j.compag.2022.106983}.
\bibitem[{Nash et~al.(1994)Nash, Sellers, Talbot, Cawthorn and
  Ford}]{nash1994population}
\bibinfo{author}{Nash, W.J.}, \bibinfo{author}{Sellers, T.L.},
  \bibinfo{author}{Talbot, S.R.}, \bibinfo{author}{Cawthorn, A.J.},
  \bibinfo{author}{Ford, W.B.}, \bibinfo{year}{1994}.
\newblock \bibinfo{title}{The population biology of abalone (haliotis species)
  in {T}asmania: blacklip abalone ({H}. rubra) from the north coast and islands
  of bass strait}.
\newblock \bibinfo{journal}{Sea Fisheries Division, Technical Report}
  \bibinfo{volume}{48}, \bibinfo{pages}{p411}.
\bibitem[{Ollila et~al.(2012)Ollila, Tyler, Koivunen and Poor}]{Ollila_2012}
\bibinfo{author}{Ollila, E.}, \bibinfo{author}{Tyler, D.E.},
  \bibinfo{author}{Koivunen, V.}, \bibinfo{author}{Poor, H.V.},
  \bibinfo{year}{2012}.
\newblock \bibinfo{title}{Complex elliptically symmetric distributions: Survey,
  new results and applications}.
\newblock \bibinfo{journal}{IEEE Trans. Signal Process.} \bibinfo{volume}{60},
  \bibinfo{pages}{5597--5625}.
\newblock \DOIprefix\doi{10.1109/TSP.2012.2212433}.
\bibitem[{Pascal et~al.(2013)Pascal, Bombrun, Tourneret and
  Berthoumieu}]{Pascal_2013}
\bibinfo{author}{Pascal, F.}, \bibinfo{author}{Bombrun, L.},
  \bibinfo{author}{Tourneret, J.Y.}, \bibinfo{author}{Berthoumieu, Y.},
  \bibinfo{year}{2013}.
\newblock \bibinfo{title}{Parameter estimation for multivariate generalized
  {G}aussian distributions}.
\newblock \bibinfo{journal}{IEEE Trans. Signal Process.} \bibinfo{volume}{61},
  \bibinfo{pages}{5960--5971}.
\newblock \DOIprefix\doi{10.1109/TSP.2013.2282909}.
\bibitem[{Pedregosa et~al.(2011)Pedregosa, Varoquaux, Gramfort, Michel,
  Thirion, Grisel, Blondel, Prettenhofer, Weiss, Dubourg, Vanderplas, Passos,
  Cournapeau, Brucher, Perrot and Duchesnay}]{scikit-learn}
\bibinfo{author}{Pedregosa, F.}, \bibinfo{author}{Varoquaux, G.},
  \bibinfo{author}{Gramfort, A.}, \bibinfo{author}{Michel, V.},
  \bibinfo{author}{Thirion, B.}, \bibinfo{author}{Grisel, O.},
  \bibinfo{author}{Blondel, M.}, \bibinfo{author}{Prettenhofer, P.},
  \bibinfo{author}{Weiss, R.}, \bibinfo{author}{Dubourg, V.},
  \bibinfo{author}{Vanderplas, J.}, \bibinfo{author}{Passos, A.},
  \bibinfo{author}{Cournapeau, D.}, \bibinfo{author}{Brucher, M.},
  \bibinfo{author}{Perrot, M.}, \bibinfo{author}{Duchesnay, E.},
  \bibinfo{year}{2011}.
\newblock \bibinfo{title}{Scikit-learn: Machine learning in {P}ython}.
\newblock \bibinfo{journal}{J. Mach. Learn. Res.} \bibinfo{volume}{12},
  \bibinfo{pages}{2825--2830}.
\newblock \URLprefix \url{http://jmlr.org/papers/v12/pedregosa11a.html}.
\bibitem[{Peel and McLachlan(2000)}]{Peel_2000}
\bibinfo{author}{Peel, D.}, \bibinfo{author}{McLachlan, G.J.},
  \bibinfo{year}{2000}.
\newblock \bibinfo{title}{Robust mixture modelling using the t distribution}.
\newblock \bibinfo{journal}{Stat. Comput.} \bibinfo{volume}{25},
  \bibinfo{pages}{339--348}.
\newblock \DOIprefix\doi{https://doi.org/10.1023/A:1008981510081}.
\bibitem[{Roizman et~al.(2020)Roizman, Jonckheere and
  Pascal}]{roizman2020flexible}
\bibinfo{author}{Roizman, V.}, \bibinfo{author}{Jonckheere, M.},
  \bibinfo{author}{Pascal, F.}, \bibinfo{year}{2020}.
\newblock \bibinfo{title}{A flexible {EM}-like clustering algorithm for noisy
  data}.
\newblock \bibinfo{journal}{To appear}
  \href{http://arxiv.org/abs/1907.01660}{\tt arXiv:1907.01660}.
\bibitem[{Roizman et~al.(2021)Roizman, Jonckheere and
  Pascal}]{roizman2021_eusipco}
\bibinfo{author}{Roizman, V.}, \bibinfo{author}{Jonckheere, M.},
  \bibinfo{author}{Pascal, F.}, \bibinfo{year}{2021}.
\newblock \bibinfo{title}{Robust clustering and outlier rejection using the
  {M}ahalanobis distance distribution}, in: \bibinfo{booktitle}{Proc. European
  Signal Processing Conference ({EUSIPCO})}, \bibinfo{address}{Amsterdam, NL}.
  pp. \bibinfo{pages}{2448--2452}.
\newblock \DOIprefix\doi{10.23919/Eusipco47968.2020.9287356}.
\bibitem[{Ruan et~al.(2011)Ruan, Yuan and Zou}]{ruan_2011}
\bibinfo{author}{Ruan, L.}, \bibinfo{author}{Yuan, M.}, \bibinfo{author}{Zou,
  H.}, \bibinfo{year}{2011}.
\newblock \bibinfo{title}{Regularized parameter estimation in high-dimensional
  gaussian mixture models}.
\newblock \bibinfo{journal}{Neural. Comput.} \bibinfo{volume}{23},
  \bibinfo{pages}{1605--1622}.
\newblock \DOIprefix\doi{10.1162/NECO_a_00128}.
\bibitem[{Salberg(2011)}]{salberg2011}
\bibinfo{author}{Salberg, A.B.}, \bibinfo{year}{2011}.
\newblock \bibinfo{title}{Land cover classification of cloud-contaminated
  multitemporal high-resolution images}.
\newblock \bibinfo{journal}{IEEE Transactions on Geoscience and Remote Sensing}
  \bibinfo{volume}{49}, \bibinfo{pages}{377--387}.
\newblock \DOIprefix\doi{10.1109/TGRS.2010.2052464}.
\bibitem[{Shen et~al.(2015)Shen, Li, Cheng, Zeng, Yang, Li and
  Zhang}]{Shen2015}
\bibinfo{author}{Shen, H.}, \bibinfo{author}{Li, X.}, \bibinfo{author}{Cheng,
  Q.}, \bibinfo{author}{Zeng, C.}, \bibinfo{author}{Yang, G.},
  \bibinfo{author}{Li, H.}, \bibinfo{author}{Zhang, L.}, \bibinfo{year}{2015}.
\newblock \bibinfo{title}{Missing information reconstruction of remote sensing
  data: A technical review}.
\newblock \bibinfo{journal}{IEEE Geosci.Remote Sens. Mag.} \bibinfo{volume}{3},
  \bibinfo{pages}{61--85}.
\newblock \DOIprefix\doi{10.1109/MGRS.2015.2441912}.
\bibitem[{Sportisse et~al.(2021)Sportisse, Biernacki, Boyer, Josse, Lourdelle,
  Celeux and Laporte}]{sportisse2021modelbased}
\bibinfo{author}{Sportisse, A.}, \bibinfo{author}{Biernacki, C.},
  \bibinfo{author}{Boyer, C.}, \bibinfo{author}{Josse, J.},
  \bibinfo{author}{Lourdelle, M.M.}, \bibinfo{author}{Celeux, G.},
  \bibinfo{author}{Laporte, F.}, \bibinfo{year}{2021}.
\newblock \bibinfo{title}{Model-based clustering with missing not at random
  data}.
\newblock \bibinfo{journal}{To appear}
  \href{http://arxiv.org/abs/2112.10425}{\tt arXiv:2112.10425}.
\bibitem[{Stekhoven and Buhlmann(2011)}]{Stekhoven_missforest_2011}
\bibinfo{author}{Stekhoven, D.J.}, \bibinfo{author}{Buhlmann, P.},
  \bibinfo{year}{2011}.
\newblock \bibinfo{title}{{Miss{F}orest. {N}non-parametric missing value
  imputation for mixed-type data}}.
\newblock \bibinfo{journal}{Bioinformatics} \bibinfo{volume}{28},
  \bibinfo{pages}{112--118}.
\newblock \DOIprefix\doi{10.1093/bioinformatics/btr597}.
\bibitem[{Tadjudin and Landgrebe(2000)}]{Tadjudin_2000}
\bibinfo{author}{Tadjudin, S.}, \bibinfo{author}{Landgrebe, D.},
  \bibinfo{year}{2000}.
\newblock \bibinfo{title}{Robust parameter estimation for mixture model}.
\newblock \bibinfo{journal}{IEEE Trans. Geosci. Remote Sens.}
  \bibinfo{volume}{38}, \bibinfo{pages}{439--445}.
\newblock \DOIprefix\doi{10.1109/36.823939}.
\bibitem[{Troyanskaya et~al.(2001)Troyanskaya, Cantor, Sherlock, Brown, Hastie,
  Tibshirani, Botstein and Altman}]{Troyanskaya2001}
\bibinfo{author}{Troyanskaya, O.}, \bibinfo{author}{Cantor, M.},
  \bibinfo{author}{Sherlock, G.}, \bibinfo{author}{Brown, P.},
  \bibinfo{author}{Hastie, T.}, \bibinfo{author}{Tibshirani, R.},
  \bibinfo{author}{Botstein, D.}, \bibinfo{author}{Altman, R.B.},
  \bibinfo{year}{2001}.
\newblock \bibinfo{title}{{Missing value estimation methods for {DNA}
  microarrays}}.
\newblock \bibinfo{journal}{Bioinformatics} \bibinfo{volume}{17},
  \bibinfo{pages}{520--525}.
\newblock \DOIprefix\doi{10.1093/bioinformatics/17.6.520}.
\bibitem[{Wang et~al.(2004)Wang, Zhang, Luo and Wei}]{WANG2004701}
\bibinfo{author}{Wang, H.X.}, \bibinfo{author}{Zhang, Q.B.},
  \bibinfo{author}{Luo, B.}, \bibinfo{author}{Wei, S.}, \bibinfo{year}{2004}.
\newblock \bibinfo{title}{Robust mixture modelling using multivariate
  t-distribution with missing information}.
\newblock \bibinfo{journal}{Pattern Recognition Lett.} \bibinfo{volume}{25},
  \bibinfo{pages}{701--710}.
\newblock \DOIprefix\doi{https://doi.org/10.1016/j.patrec.2004.01.010}.
\bibitem[{Wei et~al.(2019)Wei, Tang and McNicholas}]{WEI201918}
\bibinfo{author}{Wei, Y.}, \bibinfo{author}{Tang, Y.},
  \bibinfo{author}{McNicholas, P.D.}, \bibinfo{year}{2019}.
\newblock \bibinfo{title}{Mixtures of generalized hyperbolic distributions and
  mixtures of skew-t distributions for model-based clustering with incomplete
  data}.
\newblock \bibinfo{journal}{Comput. Stat. Data. Anal.} \bibinfo{volume}{130},
  \bibinfo{pages}{18--41}.
\newblock \DOIprefix\doi{https://doi.org/10.1016/j.csda.2018.08.016}.

\end{thebibliography}


\end{document}